%% file: iclr2021_conference.tex
\documentclass{article} 
\usepackage{times}
\usepackage{xcolor}

\usepackage[sort&compress,numbers]{natbib}

\input{math_commands.tex}

\usepackage{hyperref}
\usepackage{url}
\usepackage{amssymb}
\usepackage[ruled,vlined]{algorithm2e}
\usepackage{algpseudocode}
\usepackage[section]{placeins}
\usepackage{authblk}

\usepackage{graphicx}

\usepackage{tabularx}
\usepackage{multirow}

\newcommand{\var}{\Var}     
\newcommand{\cov}{\Cov}

\newcommand\sample[2]{
	{#1}_{#2}
}

\newtheorem{theorem}{Theorem}
\title{Gradients should stay on Path: Better Estimators of the Reverse- and Forward KL Divergence for Normalizing Flows}

\author[1]{\small Lorenz Vaitl}
\author[1, 2]{\small Kim A. Nicoli}
\author[1, 2, 3]{\small Shinichi Nakajima}
\author[1, 2]{\small Pan Kessel}

\affil[1]{\footnotesize Machine Learning Group, Department of Electrical Engineering \& Computer Science, Technische Universit\"at Berlin, Germany}
\affil[2]{\footnotesize BIFOLD - Berlin Institute for the Foundations of Learning and Data, Technische Universit\"at Berlin, Berlin, Germany}
\affil[3]{\footnotesize RIKEN Center for AIP, 103-0027 Tokyo, Chuo City, Japan}

\begin{document}

\maketitle

\begin{abstract}
We propose an algorithm to estimate the path-gradient of both the reverse and forward Kullback--Leibler divergence for an arbitrary manifestly invertible normalizing flow. The resulting path-gradient estimators are straightforward to implement, have lower variance, and lead not only to faster convergence of training but also to better overall approximation results compared to standard total gradient estimators. We also demonstrate that path-gradient training is less susceptible to mode-collapse. In light of our results, we expect that path-gradient estimators will become the new standard method to train normalizing flows for variational inference.
\end{abstract}

\section{Introduction}
Many important physical systems can be described by a Boltzmann distribution 
\begin{align}
 p(x)=\frac{1}{Z} \exp(-S(x)) \,,
\end{align}
where $S$ is the action which is often known in closed form and $Z=\int \textrm{d}^dx \, \exp(-S(x))$ denotes the partition function. The partition function is typically intractable, i.e. cannot be calculated as it is a very high-dimensional integral. Nevertheless, well-established Monte-Carlo-Markov-Chain (MCMC) can be used to sample from the target $p$ and allow for the estimation of physical observables. However, MCMC methods become extremely expensive for situations in which subsequent samples of the Markov Chain have large autocorrelation. Such critical slowing down arises for many systems of great physical interest, e.g. for critical phenomena in statistical physics, in the continuum limit of lattice field theories, or for atomistic systems with a large number of local free energy minima in context of quantum chemistry. As a result, overcoming critical slowing down constitutes one of the most important unsolved problems of modern computational physics.

Recent work \citep{noe2019boltzmann, wu2020stochastic, albergo2019flow, kanwar2020equivariant, boyda2021sampling, wu2019solving, nicoli2021estimation,  nicoli2020asymptotically, nicoli2019comment} has proposed to combine generative models with MCMC to overcome critical slowing down. In this context, a particularly promising type of generative model are normalizing flows because they allow for one-shot sampling, provide a normalized density and can be interpreted as a diffeomorphic field redefinition of the underlying physical degrees of freedom. In this approach, a normalizing flow $q$ is first trained to closely approximate a target density $p$. Afterwards, physical observables can be estimated with the same asymptotic guarantees as for established MCMC methods \citep{noe2019boltzmann, muller2019neural, nicoli2020asymptotically}. This can be achieved by using the flow either for importance sampling or as the proposal density in a Markov Chain. If the target has been learnt well, the resulting estimate will not suffer from critical slowing down as samples are drawn (almost) independently. Flow-based methods therefore allow us to completely avoid critical slowing down provided that we can train the model $q$ well. 

Unfortunately, training of normalizing flows  represents a major challenge for many physical systems of practical relevance. Current training schemes often minimize the reverse Kullback--Leibler divergence $\textrm{KL}(q, p)$ from the normalizing flow model $q$ to the target density $p$ and show a drastic deterioration in approximation quality with growing system size or as a critical point is approached. Furthermore, training often results in mode-collapse, i.e. the flow $q$ may assign vanishing probability mass to (at least one) of the modes of the target density $p$. Mode-collapse must be avoided as it invalidates asymptotic guarantees, i.e. even in the limit of infinitely many samples the estimates of physical observables are biased. 

In this work, we propose a plug and play modification of the training procedure which alleviates its aforementioned shortcomings and works for any (manifestly invertible) normalizing flow. Specifically, we propose an algorithm to estimate the path gradient of the reverse KL divergence for normalizing flows. Unlike the conventionally used total gradient, the path gradient only takes into account the implicit dependency on the flow's parameters through reparameterized sampling but is insensitive to any explicit dependency. We demonstrate that the resulting path gradient estimator has lower variance compared to the standard estimator and leads to faster convergence of training as a result.   

Furthermore, we demonstrate that a path-gradient estimator can also be used to minimize the forward Kullback--Leibler divergence $\textrm{KL}(p, q)$ which is known to be significantly more robust to mode-collapse, see e.g. \cite{hackett2021flow, nicoli2021machine, nicoli2021machine}, and therefore is the preferable choice to preserve asymptotic guarantees.

We show in detailed numerical experiments that our path-gradient method leads to superior training results and is able to significantly alleviate mode dropping. We also study these path gradient estimators theoretically by analyzing their statistical properties in various phases of the training process.

\subsection{Related Works}
Broadly speaking, our work builds on and significantly extends \citet{roeder2017sticking} and \citet{vaitl2022path} which propose path gradient estimators which only work for simple Gaussian variational models or for the very restricted subclass of continuous normalizing flows, respectively.

More specifically, \citet{roeder2017sticking} proposed a path-gradient estimator for the case of a Gaussian variational density and the standard ELBO loss in the context of variational autoencoders (VAE). In \citet{tucker2018doubly}, the authors extended these results to other VAE losses, such as Importance Weighted Autoencoder \citep{burda2015importance}, Reweighted Wake Sleep \citep{bornschein2014reweighted}, and Jackknife Variational Autoencoder \citep{nowozin2018debiasing}. More specifically, the authors proposed an identity which allows to rewrite any REINFORCE-based estimator\citep{williams1992simple} as a path-wise gradient estimator. Both references empirically demonstrated superior performance of the path-wise estimators for VAEs. Later work \citep{finke2019importanceweighted, bauer2021generalized, geffner2020difficulty, geffner2021empirical} extended these results to other VAE loss functions, for example based on $\alpha$-divergences, and clarified theoretical aspects of the original references. It is important to stress that all the aforementioned works require a simple variational model, such as a Gaussian.

Our work focuses on normalizing flows which allow to model complex target distributions of physical systems -- in stark contract to simple Gaussian variational densities. Standard approaches \cite{noe2019boltzmann, muller2019neural} minimize the reverse and forward Kullback--Leibler divergence using the standard total gradient, as opposed to the path gradient, estimator.  

To the best of our knowledge, \citet{agrawal2020advances} is the only reference studying path-wise gradient estimators of general invertible normalizing flows. This study is however limited to the standard reverse Kullback--Leibler divergence as part of a broader ablation for comparatively simple models from the STAN library. In contrast to our contribution, their study does not consider modelling complex distributions of physical systems and path-gradients of forward KL losses. Furthermore, their proposed estimation algorithm has twice the memory costs, severely limiting its suitability for physics applications, as opposed to our proposal. 

More recently, \citet{vaitl2022path} introduced an efficient path-gradient estimator for continuous normalizing flows (CNFs). This algorithm for estimating the path gradient -- even though efficient -- is tailored to CNFs and requires rewriting the gradient computation. Our approach on the other hand is applicable to an arbitrary manifestly invertible normalizing flow architecture. In particular, it is also applicable to the case of a continuous normalizing flow. For this specific case however, it is less efficient than the method proposed in \cite{vaitl2022path}. 

Therefore, the present work can be thought of as generalization of \cite{vaitl2022path} to any manifestly invertible normalizing flow architectures. A further notable novelty of our work is that we discuss estimation of the path gradient of the forward KL for normalizing flows.


\subsection{Sampling with Normalizing Flows}\label{sec:samplingwithflows}
\begin{figure}[t]
    \centering
    \includegraphics[width=\textwidth]{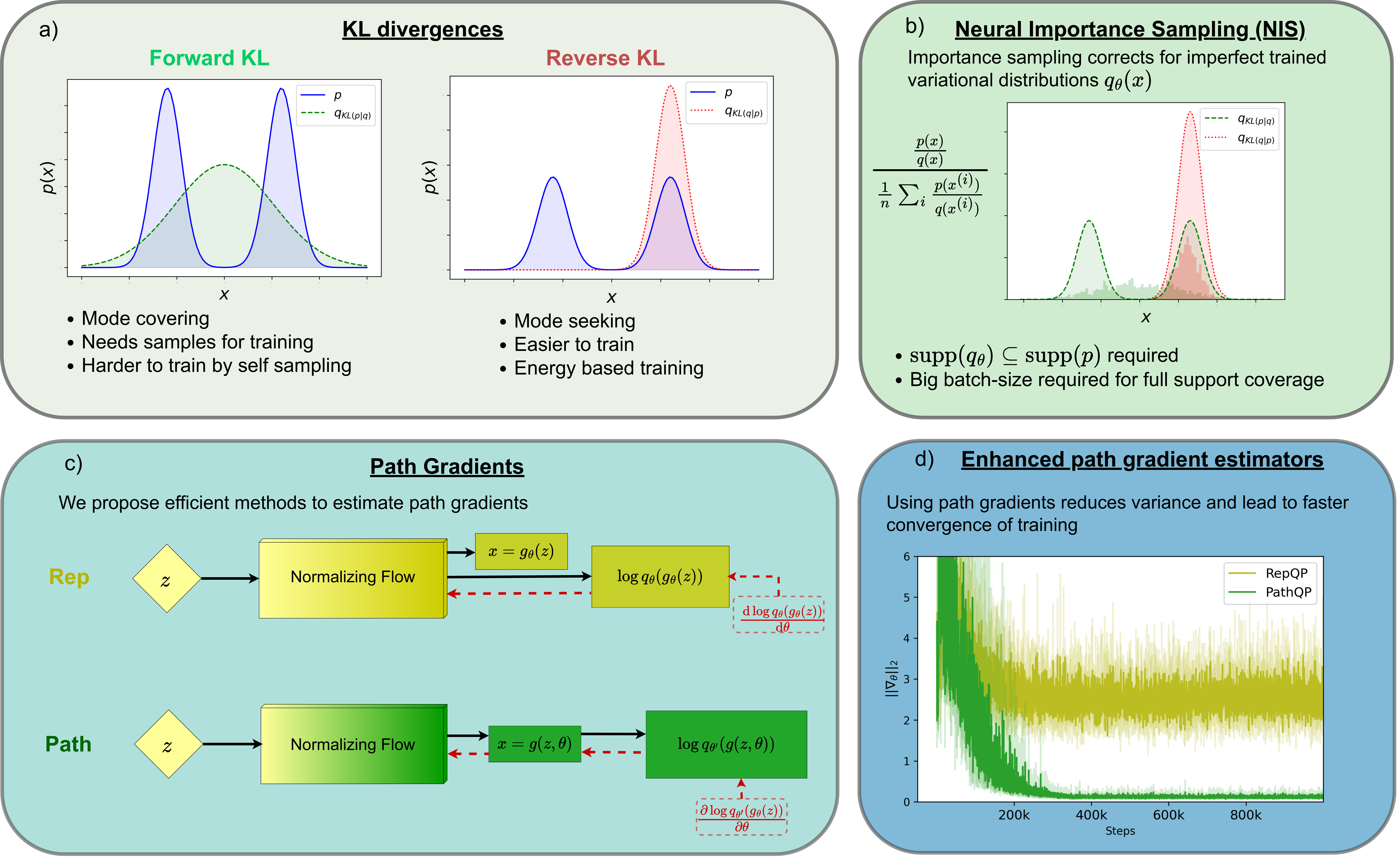}
    \caption{summary of the paper. \textbf{a)} Forward and reverse KL divergences being respectively \textit{mode covering} and \textit{mode seeking}. The latter is thus more prone to \textit{mode dropping}. \textbf{b)} Neural importance sampling (NIS) scheme corrects for imperfections of the learned variational distribution, when the support of $p(x)$ is covered. \textbf{c)} We propose to use path gradients and develop an efficient path gradient estimator to improve training efficiency through faster convergence and lower variance. \textbf{d)} We empirically demonstrate that using path gradients (PathQP) over standard total gradient estimators (RepQP) substantially improve the performances.}
    \label{fig:emp-stl-gradnorm}
\end{figure}
A normalizing flow is a bijective map $g_\theta: \mathcal{Z} \to \mathcal{X}$ from a base space $\mathcal{Z} \subset \mathbb{R}^d$ to a target space $\mathcal{X} \subset \mathbb{R}^d$. The base space $\mathcal{Z}$ is equipped with a simple probability density $q_Z$. The bijection $g_\theta$ then induces a probability density $q_\theta$ on the target space $\mathcal{X}$ by
\begin{align}
    q_{\theta} (x) = q_Z (g_\theta^{-1}(x)) \left|\frac{\partial g_\theta^{-1}(x)}{\partial x}\right| \,, \label{eq:logprob}
\end{align}
where $\left|\frac{\partial g_\theta^{-1}(x)}{\partial x}\right|$ denotes the absolute value of the determinant of the Jacobian of the inverse flow $g_\theta^{-1}$. 

The flow is trained to closely approximate the target density $p$. As we will explain in detail in the next section, this can be done even for an intractable partition function. 

After training, we can use the flow $q_\theta$ to sample from the target density $p$ with asymptotic guarantees. To this end, one uses neural importance sampling \citep{noe2019boltzmann, muller2019neural, nicoli2020asymptotically} to estimate the expectation value of some observable $\mathcal{Q}$ with respect to the target density $p$ by
\begin{align}
    \mathbb{E}_{x \sim p} \left[ \mathcal{Q}(x) \right] &= \mathbb{E}_{x \sim q_\theta} \left[ w(x) \, \mathcal{Q}(x) \right] 
    \approx \frac{1}{N} \sum_{i=1}^N \hat w_i \, \mathcal{Q}(x_i) \,, && \textrm{with}  \; \; x_i \sim q_\theta \,, \label{eq:iss}
\end{align}
where we have defined the normalized importance weight 
$$
w(x) = \frac{p(x)}{q(x)}\, ,
$$
and its estimator $\hat w$ as
\begin{align}
    \hat w_i = \hat w(x_i) = \frac{1}{\hat{Z}} \frac{\exp(-S(x_i))}{q_\theta(x_i)} \,,
\end{align}
which uses the estimator $\hat{Z}$ of the partition function\footnote{We suppress the dependence of the estimator $\hat{Z}$ on the number of samples $N$ to alleviate notation.} 
\begin{align*}
Z = \int \textrm{d}^d x \, q_\theta(x) \frac{\exp(-S(x))}{q_\theta(x)} && \approx && \hat{Z} = \frac{1}{N} \sum_{i=1}^N \frac{\exp(-S(x_i))}{q_\theta(x)} \,, && x_i \sim q_\theta \,.
\end{align*}
The variance of the estimator \eqref{eq:iss} is given by
\begin{align*}
    \sigma^2 = \textrm{Var}(\mathcal{Q}) \frac{1}{N \,\textrm{ESS}} + o_p(N^{-1})
\end{align*}
with the effective sampling size
\begin{align}
    \textrm{ESS} = \frac{1}{\mathbb{E}_q [ w(x) ]} \in [0, 1] \,. \label{eq:ess}
\end{align}
The effective sampling size is one for a perfectly trained sampler while very low for a poorly trained one. The effective sampling size therefore provides a natural metric to quantify the quality of a sampler.

If the model density $q$ has larger or equal support than the target density, i.e.
\begin{align}
    \textrm{supp}(p) \subseteq \textrm{supp}(q_\theta) \,, \label{eq:largersupp}
\end{align}
normalized importance sampling \eqref{eq:iss} provides a statistically consistent estimator of the expectation value $\mathbb{E}_p [ \mathcal{O} ]$ and thus has the same asymptotic guarantees as well-established Monte-Carlo-Markov-Chain (MCMC) methods. We refer to \citet{nicoli2020asymptotically} for a detailed proof.
In contrast to popular MCMC algorithms, this method uses independent and identically distributed (iid) samples from the flow. Flow-based sampling may therefore have considerable advantages over established Markov chain techniques for situations in which MCMC suffers from large autocorrelation or has problems overcoming large barriers in the action landscape.

We also mention in passing that flow-based sampling can be combined with MCMC by using the flow as the proposal density of the Markov chain \citep{noe2019boltzmann, nicoli2020asymptotically, albergo2019flow}. The resulting algorithm is called Neural MCMC. The proposal is drawn independently and therefore does not depend on the previous element of the chain. This is in stark contrast to conventional MCMC algorithms which rely on a (typcially small) random modification of the previous configuration to create a proposal. Crucially, Neural MCMC also only comes with asymptotic guarantees if the flow has larger support than the target, i.e. provided that \eqref{eq:largersupp} holds. 

As we will discuss, current training approaches often lead to a violation of the larger support requirement \eqref{eq:largersupp} due to mode-collapse. One of the central motivations of the present work is to propose modifications of the training procedure to alleviate this effect.

\subsection{Training with Reverse KL}\label{sec:revtraining}
The flow can be trained by minimizing the reverse KL divergence
\begin{align}
    \textrm{KL}(q_\theta, p) = \mathbb{E}_{x \sim q_\theta} \left[ S(x) + \log q_\theta(x) \right] + \textrm{const.} \,,
\end{align}
where the last summand denotes terms independent of the parameters $\theta$ of the flow $q_\theta$. As a result, this term will have no contribution to the gradient of the loss function. This gradient can be rewritten using the reparametrization trick, i.e. $\mathbb{E}_{x \sim q_\theta}[f(x)] = \mathbb{E}_{z \sim q_z}[f(g_\theta(z))]$, and is then given by 
\begin{align}
    \frac{d}{d\theta}  \textrm{KL}(q_\theta, p) = \mathbb{E}_{z \sim q_Z} \left[ \frac{d}{d\theta}S(g_\theta(z))+ \frac{d}{d\theta} \log q_\theta(g_\theta(z)) \right] \,. \label{eq:gradKL}
\end{align}
It is straightforward to obtain a Monte-Carlo estimator of this gradient,
\begin{align}
\frac{d}{d\theta}  \textrm{KL}(q_\theta, p) \approx \mathcal{G}_{\textrm{RepQP}},
\end{align}
by drawing samples from the base density $q_Z$, reparametrizing and calculating the gradients of the action and of the log-probability by backpropagation, i.e.
\begin{align}
\mathcal{G}_{\textrm{RepQP}} = \frac{1}{N} \sum_{i=1}^{N} \left( \frac{d}{d\theta}S(g_\theta(z_i))+ \frac{d}{d\theta} \log q_\theta(g_\theta(z_i)) \right) \,,  && z_i \sim q_Z \,. \label{eq:reparamRev}
\end{align}
We will refer to $\mathcal{G}_{\textrm{RepQP}}$ as the \emph{reparameterized $qp$-estimator (RepQP)}. Currently, this estimator is the most widely used.

However, the RepQP estimator can often be suboptimal and we propose to instead use the \emph{path-gradient estimator}. For this, it is convenient to define the path-gradient of an arbitrary function $f(g_\theta(z), \theta)$  by
\begin{align*}
   \blacktriangledown_\theta f(g_\theta(z), \theta) = \frac{\partial f(g_\theta(z), \theta)}{\partial g_{\theta}(z)}\frac{\partial g_{\theta}(z)}{\partial \theta} \,,
\end{align*}
which implies that its total derivative can be written as
\begin{align}
    \frac{d}{d\theta} f(g_\theta(z),\theta) = \blacktriangledown_\theta f(g_\theta(z),\theta)+ \left. \frac{\partial}{\partial \theta} f(x, \theta) \right|_{x=g_\theta(z)} \,, \label{eq:totalderpath}
\end{align}
i.e. the path derivative only takes into account the implicit dependency on $\theta$ through the flow $g_{\theta}$ and is insensitive to any explicit dependency. Using this definition, we can rewrite the gradient \eqref{eq:gradKL} of the KL-divergence as
\begin{align*}
     \frac{d}{d\theta}  \textrm{KL}(q_\theta, p) = \mathbb{E}_{z \sim q_Z} \left[ \blacktriangledown_\theta S(g_\theta(z))+ \blacktriangledown_\theta \log q_\theta(g_\theta(z)) \right]  +  \mathbb{E}_{z \sim q_Z} \left[ \left. \frac{\partial}{\partial \theta} \log q_\theta(x) \right|_{x=g_\theta(z)} \right] \,,
\end{align*}
where we have used that the path and total gradient lead to the same result for the action as
\begin{align}
    \frac{d}{d \theta} S(g_\theta(z)) = \frac{\partial S(g_\theta(z))}{\partial g_{\theta}(z)}\frac{\partial g_{\theta}(z)}{\partial \theta} = \blacktriangledown_\theta S(g_\theta(z)) \,. \label{eq:pathtotalactioneq}
\end{align}
By applying the reparameterization trick again, it is easy to see that the last \emph{score term} vanishes
\begin{align}
    \mathbb{E}_{z \sim q_Z} \left[ \left. \frac{\partial}{\partial \theta} \log q_\theta(x) \right|_{x=g_\theta(z)} \right] = \mathbb{E}_{x \sim q_\theta} \left[ \frac{\partial}{\partial \theta} \log q_\theta(x) \right] =   
    \frac{\partial}{\partial \theta} \int \textrm{d}^dx \, q_\theta(x) = 0  \,. \label{eq:scoreTerm}
\end{align}
By explicitly excluding the vanishing score term from the gradient of the KL-divergence, we obtain the \emph{path-gradient $qp$-estimator (PathQP)}  
\begin{align}
    \frac{d}{d\theta}  \textrm{KL}(q_\theta, p) \approx \mathcal{G}_{\textrm{PathQP}}\,,
\end{align}    
which is given by
\begin{align}
    \mathcal{G}_{\textrm{PathQP}} = \frac{1}{N} \sum_{i=1}^N \left( \blacktriangledown_\theta S(g_\theta(z_i))+ \blacktriangledown_\theta \log q_\theta(g_\theta(z_i))  \right) &&\, z_i \sim q_Z\,. \label{eq:pathqp}
\end{align}
Both the path-gradient and the reparameterized $qp$-estimator are unbiased estimators of the gradient of the reverse KL divergence. However, their variances are generically different. This effect is particularly pronounced if the variational distribution perfectly approximates the target, i.e.
\begin{align*}
    \forall x \in \mathcal{X}: && q_\theta(x) = p(x) \,.
\end{align*}
For such a perfect approximation, the path-gradient estimator vanishes identically:
\begin{align}
    \mathcal{G}_{\textrm{PathQP}} &=\frac{1}{N} \sum_{i=1}^N \left( \blacktriangledown_\theta S(g_\theta(z_i))+ \blacktriangledown_\theta \log q_\theta(g_\theta(z_i))  \right) \nonumber \\
    &=  -\frac{1}{N} \sum_{i=1}^N  \blacktriangledown_\theta \log \underbrace{\left( \frac{p(g_\theta(z_i)}{q_\theta(g_\theta(z_i))} \right) }_{=1} = 0 \,, \label{eq:vanishingvar}
\end{align}
where we have used that $- \blacktriangledown_\theta \log p(g_\theta(z)) = \blacktriangledown_\theta S(g_\theta(z))$.  As a result, the variance of the path gradient estimator $\mathcal{G}_{\textrm{PathQP}}$ vanishes in this limit. This is in contrast to the reparameterized estimator which can be rewritten as
\begin{align}
    \mathcal{G}_{\textrm{RepQP}} = \mathcal{G}_{\textrm{PathQP}} + \mathcal{G}_{\textrm{Score}}
\end{align}
where we have defined
\begin{align}
    \mathcal{G}_{\textrm{Score}} = \frac{1}{N} \sum_{i=1}^N  \frac{\partial}{\partial \theta} \log q_\theta (g_\theta(z_i)) \,,  && z_i \sim q_Z \,,
\end{align}
whose variance is given by $\textrm{Var}[\mathcal{G}_{\textrm{Score}}] = \frac{\mathcal{I}(\theta)}{N}$ where we have defined the Fisher information
\begin{align}
\mathcal{I}(\theta)=\mathbb{E}_{x \sim q_\theta} \left[ \frac{\partial}{\partial \theta} \log q_\theta(x) \frac{\partial}{\partial \theta} \log q_\theta(x)\right]
\end{align}
of the variational distribution. 
As a result, the reparameterized gradient estimator $\mathcal{G}_{\textrm{RepQP}}$ has generically non-vanishing variance even if the variational distribution perfectly approximates the target distribution. 
By continuity, we may expect that the variance of the reparameterized estimator is substantially larger than the path-gradient estimator in the final phase of training and that using the path-gradient estimator will lead to a better convergence of training as a result. We will indeed demonstrate this in the numerical experiments.

\begin{figure}
    \centering
    \includegraphics[width=0.8\textwidth]{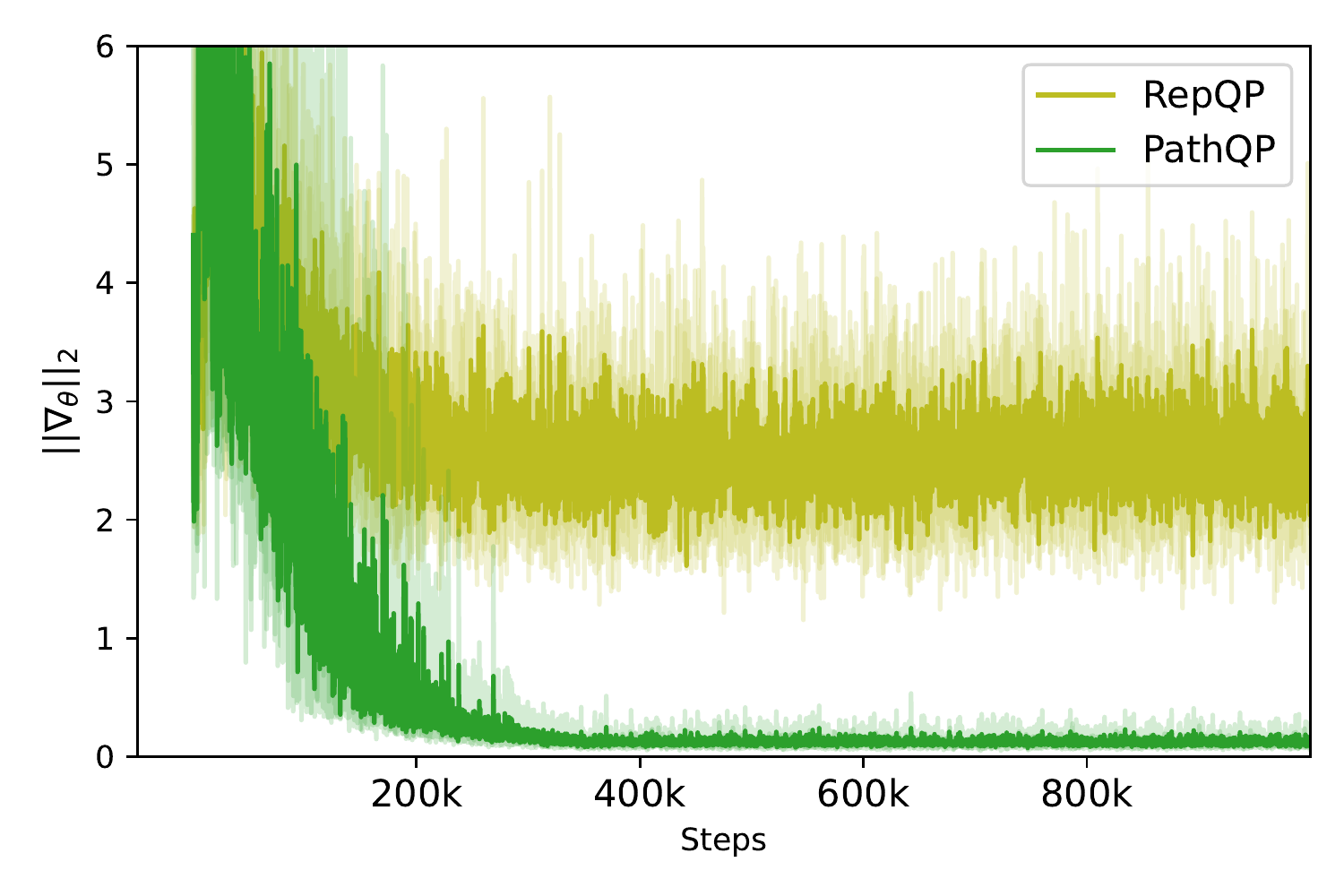}
    \caption{The norm of the gradient estimated by PathQP goes to zero when the target density $p$ is well approximated towards the end of the training. The norms of the gradients are averaged over three runs for the Double-Well experiments with 8 timesteps and $m_0=2.75$ in \Eqref{eq:dw-pot}.}
    \label{fig:emp-STL-gradnorm}
\end{figure}

\section{Training with Forward KL}\label{sec:forwardTraining}
As discussed in Section~\ref{sec:samplingwithflows}, neural importance sampling and neural Markov-Chains require the flow $q_\theta$ to have larger support than the target density $p$ in order to be statistically consistent. Training with the reverse KL divergence is therefore problematic as it can lead to mode-collapse. Furthermore, in practice, it may also be the case that the flow assigns only an infinitesimal  probability mass to modes of the target density $p$. In this case, importance sampling will not lead to reasonable results for any feasible number of samples (although strictly speaking, it is still statistically consistent in the limit of infinite number of samples). 

For this reason, it is preferable to train the flow by minimizing the forward KL divergence,
\begin{align}
\textrm{KL}(p, q_\theta) = \mathbb{E}_{x \sim p} \left[ \log \left( \frac{p(x)}{q_\theta(x)} \right) \right] \,,
\end{align}
as it heavily penalizes mode-collapse.
A natural approach to estimate the gradient of the forward KL is to re-weight the expectation value with respect to the target $p$ such that it becomes an expectation with respect to $q$ 
\begin{align}
    \frac{d}{d\theta} \textrm{KL}(p, q_\theta) 
    &=\mathbb{E}_{x \sim p} \left[ \frac{\partial}{\partial \theta} \log \left( \frac{p(x)}{q_\theta(x)} \right) \right] \nonumber \\
    &=\mathbb{E}_{x \sim q_\theta} \left[ \frac{p(x)}{q_\theta(x)} \frac{\partial}{\partial \theta} \log \left( \frac{p(x)}{q_\theta(x)} \right) \right] \label{eq:reinforce}\\
    &=\mathbb{E}_{x \sim q_\theta} \left[ \frac{\partial}{\partial \theta}  \frac{p(x)}{q_\theta(x)}  \right] \label{eq:partialforward} \,.
\end{align}
We then use the reparameterization trick to obtain
\begin{align*}
     \frac{d}{d\theta} \textrm{KL}(p, q_\theta)   
    &=\mathbb{E}_{z \sim q_Z} \left[ \left.\frac{\partial}{\partial \theta}  \frac{p(x)}{q_\theta(x)}\right|_{x=g_\theta(z)}  \right]\,.
\end{align*}
Using the relation \eqref{eq:totalderpath} of the partial derivative in \eqref{eq:partialforward} with the path and total derivative, this can be rewritten as
\begin{align*}
     \frac{d}{d\theta} \textrm{KL}(p, q_\theta)   
    &=\mathbb{E}_{z \sim q_Z} \left[ \frac{d}{d \theta}  \frac{p(g_\theta(z))}{q_\theta(g_\theta(z))} - \blacktriangledown_\theta  \frac{p(g_\theta(z))}{q_\theta(g_\theta(z))}  \right]\,.
\end{align*}
The first summand on the right-hand-side is vanishing because
\begin{align*}
    \mathbb{E}_{z \sim q_Z} \left[ \frac{d}{d \theta}  \frac{p(g_\theta(z))}{q_\theta(g_\theta(z))} \right] = 
    \frac{d}{d \theta}\mathbb{E}_{z \sim q_Z} \left[  \frac{p(g_\theta(z))}{q_\theta(g_\theta(z))} \right] = 
    \frac{d}{d \theta}\mathbb{E}_{x \sim q_\theta} \left[  \frac{p(x)}{q_\theta(x)} \right] = 0 \,, 
\end{align*}
where we have used in the last step that $\mathbb{E}_{x \sim q_\theta} \left[  \frac{p(x)}{q_\theta(x)} \right] = \int \textrm{d}^d x \, p(x) = 1$. 

In summary, we have obtained an expression of the forward KL gradient in terms of a path-derivative, i.e.
\begin{align}
     \frac{d}{d\theta} \textrm{KL}(p, q_\theta)   
    = - \mathbb{E}_{z \sim q_Z} \left[ \blacktriangledown_\theta  \frac{p(g_\theta(z))}{q_\theta(g_\theta(z))}  \right]
    \label{eq:forwardPG}
\end{align}
We note that this relation can also be derived using the so-called DReG-identity of \citet{tucker2018doubly}.
In the next section, we discuss estimators of this forward path-gradient. 

\subsection{Estimators for the Forward KL Path-Gradient}
There are two possibilities to obtain an estimator for the forward KL path gradient \eqref{eq:forwardPG}. To see why, we recall that the exact normalized weight is defined as
\begin{align}
    w(x) = \frac{1}{Z} \tilde{w}(x) \,,
\end{align}
where  $\tilde{w}=\frac{\exp(-S(x))}{q_\theta(x)}$ is the unnormalized importance weight. Using these definitions, we can rewrite the path-gradient \eqref{eq:forwardPG} as
 \begin{align*}
    \frac{d}{d \theta} \textrm{KL} (p, q_\theta) 
   = - \mathbb{E}_{z \sim q_Z} \left[ \blacktriangledown_\theta  w(g_\theta(z))  \right]
   \approx -\frac{1}{N} \sum_{i=1}^N \blacktriangledown_\theta \frac{\tilde{w}(g_\theta(z_i))}{Z} \,.
    \end{align*}
Since the partition function $Z$ is intractable, we need to estimate it with samples from $q$ by
\begin{align}
    Z \approx \hat{Z} = \frac{1}{N} \sum_{j=1}^N \tilde{w}(g_\theta(z_i)) && z_i \sim q_Z\,.
\end{align}
There are now two ways of obtaining path-gradient estimators:
\begin{itemize}
    \item We can either pull $Z$ through the path-derivative and then estimate
    \begin{align}
    \frac{d}{d \theta} \textrm{KL} (p, q_\theta) 
    &\approx \frac{-1}{\hat{Z}} \, \frac{1}{N} \sum_{i=1}^N \blacktriangledown_\theta \tilde{w}(g_\theta(z_i)) = -\sum_{i=1}^N \frac{\tilde{w}_i}{\sum_{j=1}^N \tilde{w}_j} \blacktriangledown_\theta \log(\tilde{w}_i) \label{eq:vanilla}
    \end{align}
    We will refer to this estimator as the \emph{path-gradient $pq$-estimator (PathPQ)}. 
    \item Alternatively, we can let the path derivative act on the estimated partition function
    \begin{align}
    \frac{d}{d \theta} \textrm{KL} (p, q_\theta) 
   &\approx -\frac{1}{N} \sum_{i=1}^N \blacktriangledown_\theta \frac{\tilde{w}(g_\theta(z_i))}{\hat{Z}} \nonumber \\
   &= -\sum_{i=1}^N \left( \frac{\tilde{w}_i}{\sum_{j=1}^N \tilde{w}_j} -  \frac{\tilde{w}_i^2}{(\sum_{j=1}^N \tilde{w}_j)^2}\right) \blacktriangledown_\theta \log(\tilde{w}_i) \label{eq:dreg}
    \end{align}
    We will refer to this estimator as \emph{$Z$ path-gradient $pq$-estimator ($Z$PathPQ)}.
\end{itemize}
As a baseline, we will also consider an estimator which is not based on path-gradients: 
\begin{itemize}
    \item To this end, we estimate \eqref{eq:reinforce} by
    \begin{align}
     \frac{d}{d \theta} \textrm{KL} (p, q_\theta)
     \approx \frac{1}{N} \sum_{i=1}^N \frac{\tilde{w}_i}{\hat{Z}} \frac{\partial}{\partial \theta} \log \left( \tilde{w}_i \right) \,, \label{eq:forwardReinforce}
    \end{align}
    and refer to it as \emph{reinforce $pq$ estimator (ReinfPQ)}.
\end{itemize}
We note that, to the best of our knowledge, the PathPQ and ZPathPQ estimators were first used in the context of Reweighted-Wake-Sleep (RWS) training of Variational Autoencoders by \citet{finke2019importanceweighted} and \citet{tucker2018doubly} for simple Gaussian variational distributions, respectively. 

Both the PathPQ and the ZPathPQ estimator have vanishing variance in the limit of perfect approximation, i.e. $q_\theta \equiv p$. This immediately follows by a completely analogous argument as for the reverse KL case, see \eqref{eq:vanishingvar}. Similarly, the variance of the ReinfPQ estimator is proportional to the Fisher information of the variational distribution $q_\theta$ in this limit and is thus generically non-vanishing. We refer to Supplement~\ref{app:reinf} for a proof. 

One may therefore again expect that the path gradient estimators have lower variance than the reinforce baseline in the final phase of training and will thus lead to better convergence of training. We will verify this in the numerical experiments.

\subsection{Theoretical Analysis of Path Gradient Estimators}\label{sec:theorytrainingphases}
\paragraph{Initial Training Phase:} estimating the forward KL divergence by reweighting can be challenging in the initial phase of training as the density of the flow $q_\theta$ and the target $p$ will have small overlap. In order to analyze this initial training regime theoretically, we will assume, without loss of generality, that all samples $\{x_i\}_{i=1}^{N-1}$ but one sample $x_N$ drawn from the flow $q_\theta$ will be in regions of the sampling space for which the target density is very small and flat, i.e.
\begin{align}
    p(x_i) = \mathcal{O}(\epsilon)\,, && \nabla p(x_i) = \mathcal{O}(\epsilon) \,, && q_\theta(x_i) = \mathcal{O}(1) \,, && \nabla q_\theta(x_i) = \mathcal{O}(1) \,,
\end{align}
for small $\epsilon > 0$ and $i \in \{1, \dots, N-1\}$.\footnote{While not explicit in the notation, we assume that $q(x_i)$ is not smaller than an order one number. This assumption is reasonable as $x_i$ is a sample of the flow $q(x_i)$.} Since most interesting densities in physics have an exponential fall-off around their modes, this is a reasonable assumption in practice. Under the further mild assumption that $\frac{\partial x}{\partial \theta}$ does not diverge, this implies
\begin{align}
\blacktriangledown_\theta w_i = \mathcal{O}(\epsilon)\,, && w_i = \mathcal{O}(\epsilon) \,. 
\end{align}
We then show in the Appendix~\ref{app:initial-phase} that the PathPQ estimator \eqref{eq:vanilla} is 
\begin{align}
\sum_{i=1}^N \frac{\tilde{w}_i}{\sum_{j=1}^N \tilde{w}_j} \blacktriangledown_\theta \log(\tilde{w}_i) = \frac{\blacktriangledown_\theta \tilde{w}_N}{ \tilde{w}_N} + \mathcal{O}(\epsilon)
\end{align}
while the $Z$PathPQ estimator \eqref{eq:dreg} has no order one contribution
\begin{align}
    \sum_{i=1}^N \left( \frac{\tilde{w}_i}{\sum_{j=1}^N \tilde{w}_j} -  \frac{\tilde{w}_i^2}{(\sum_{j=1}^N \tilde{w}_j)^2}\right) \blacktriangledown_\theta \log(\tilde{w}_i) = \mathcal{O}(\epsilon) \,.
\end{align}\label{eq:zpq_kim}
The $Z$PathPQ estimator \eqref{eq:dreg} can thus be expected to struggle in the initial training phase. We empirically confirm this in \Secref{sec:experiments}.

\begin{figure}
    \centering
    \includegraphics[width=0.8\textwidth]{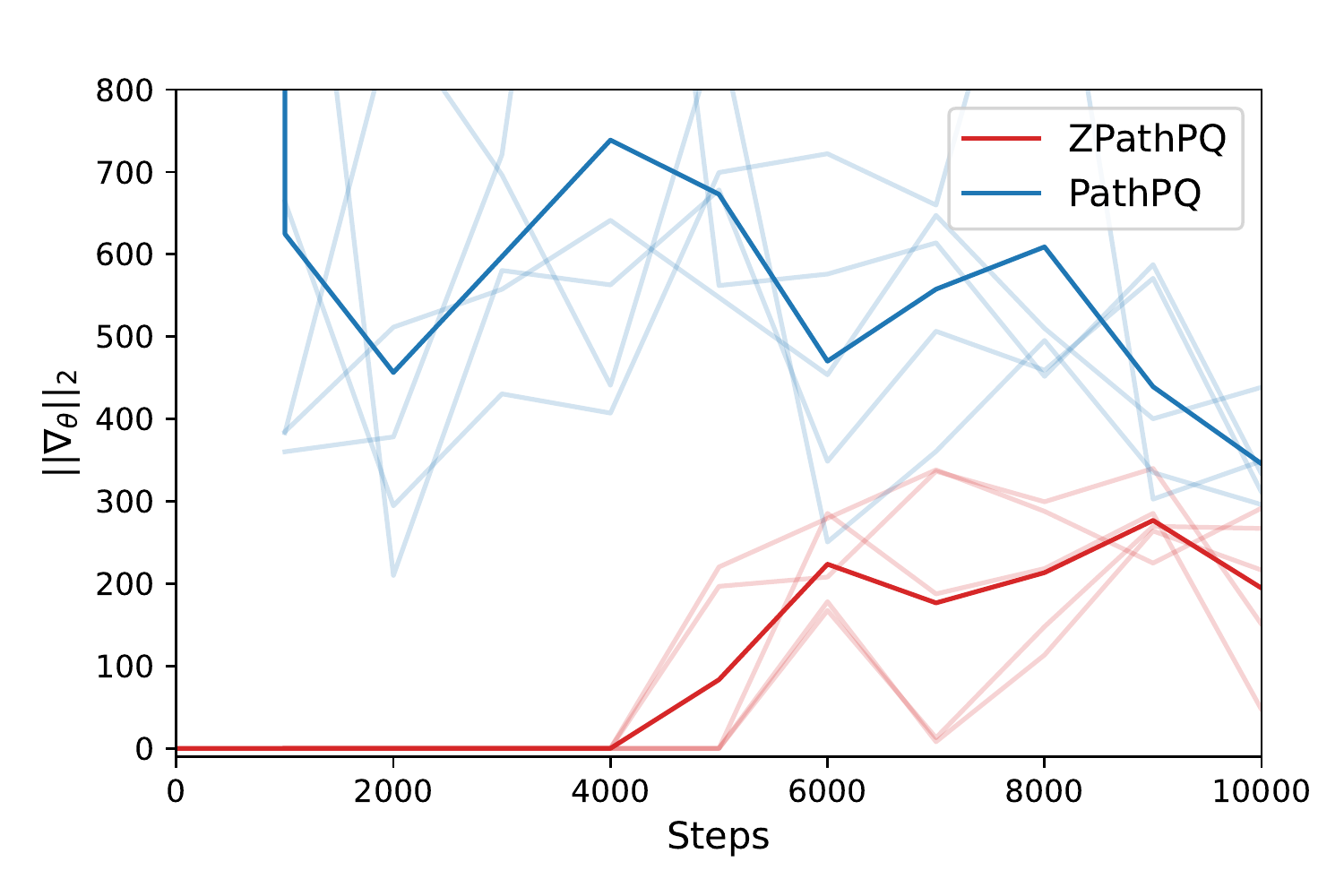}
    \caption{Norm of the $Z$PathPQ and PathPQ estimators in the inital phase of training averaged over 5 training runs for the Double-Well with 64 timesteps and $m_0=2.75$. This demonstrates that the $Z$PathPQ estimator can lead to vanishing gradients in this initial training phase - as expected from \Eqref{eq:zpq_kim}.}
    \label{fig:DReG-Initial-Phase}
\end{figure}

\paragraph{Asymptotic Training Phase:} If the flow density $q_\theta$ already approximates the target $p$ relatively well such that the variance of the normalized importance weight is small, we can use the delta method to calculate the variance and bias of the estimators. We show in the Appendix~\ref{app:Asympt-behaviour} that both the PathPQ and $Z$PathPQ estimators have the same variance and comparable bias to leading order in the number of samples $N$. Thus both estimators can be expected to lead to similar performance in this training regime.

\subsection{Implementation of Estimators}\label{sec:implementation}
In this section, we will discuss the practical implementation of the path gradient for normalizing flows. For both forward estimators \eqref{eq:vanilla} and \eqref{eq:dreg} as well as the reverse estimator, we need to calculate
\begin{align*}
    \blacktriangledown_\theta \log(\tilde{w}) = \blacktriangledown_\theta \log (q_\theta(g_\theta(z))) +  \blacktriangledown_\theta S(g_\theta(z)) \,.
\end{align*}
The second term can trivially be obtained by automatic differentiation because the total derivative leads to the same result as the path gradient, i.e. $\frac{d}{d\theta} S(g_\theta(z)) = \blacktriangledown_\theta S(g_\theta(z))$, see \eqref{eq:pathtotalactioneq}. 

The first term however does not have this property and therefore requires more care. For variational inference, the log density of a normalizing flow is typically computed along with the forward pass through the flow $x=g_\theta(z)$ which produces the sample $x$. This makes it challenging to calculate the path-derivative $\frac{\partial \log q_\theta(x)}{\partial x}$ by standard reverse-mode automatic differentiation because the sample $x$ is the output as opposed to the input of the forward pass. 

We overcome this challenge, by proposing Algorithm~\ref{alg:pathgrad} which estimates the path-gradient with the same memory footprint as needed for the conventional gradient at the cost of roughly two forward passes through the flow. In practice, a low memory footprint is crucial because invertible architectures tend to have large memory requirements which severely limits the possible batch sizes. Large batch sizes are however essential for successful training of flows by self-sampling. This is because training starts from a randomly initialized flow. If the batch size is not large enough, the probability of probing regions of the sampling space with significant probability mass tends to be low and the flow will not be able to learn to approximate the target $p$ as it has not sampled these relevant regions. This effect is particular pronounced as the system size increases as the action $S$ is an extensive quantity and the target density $p$ therefore becomes increasingly concentrated around its local minima. 

While our algorithm has approximately double runtime per gradient update compared to the standard total gradient estimator, it will be shown empirically that it leads to faster convergence of training overall as may be expected due to favorable variance properties discussed in Section~\ref{sec:revtraining}.

\begin{algorithm}
\caption{Path gradient $\blacktriangledown_\theta \log q_\theta ( g_\theta(z) )$}\label{alg:pathgrad}
\textbf{Input:} base sample $z \sim q_Z$ \\
$\;\;x' \gets \textrm{stop\_gradient}(g_\theta(z))$ {\scriptsize \hspace{4.5em} \# forward pass of $z$ through the flow without gradients}\\
$\;\; q_{\theta}(x') \gets  q_Z (g_\theta^{-1}(x')) \left|\frac{\partial g_\theta^{-1}(x')}{\partial x'}\right|$  {\scriptsize \hspace{1em} \# reverse pass to calculate density}\\
$\;\; G \gets  \frac{\partial \log(q_\theta(x'))}{\partial x'}$  {\scriptsize \hspace{10em} \# compute gradient with respect to $x'$}\\
$\;\;x \gets g_\theta(z)$ {\scriptsize\hspace{13.6em} \# standard forward pass}\\
$\;\; \textbf{return} \; \frac{d}{d \theta} \left( \textrm{stop\_gradient}(G)^T x\right)$ {\scriptsize\hspace{1em} \# for path-gradient, contract $\frac{\partial x}{\partial \theta}$ with $\partial \log q_\theta(x)/\partial x$}
\end{algorithm}

\section{Numerical Experiments}
\label{sec:experiments}
\subsection{Quantum Mechanical Particle in Double-Well}
In order to evaluate the performance of training with path-gradients, we consider a quantum mechanical particle in a double well potential which is a prototypical example for a two-moded distribution in quantum mechanics.

\begin{figure}
    \centering
    \includegraphics[width=0.48 \textwidth]{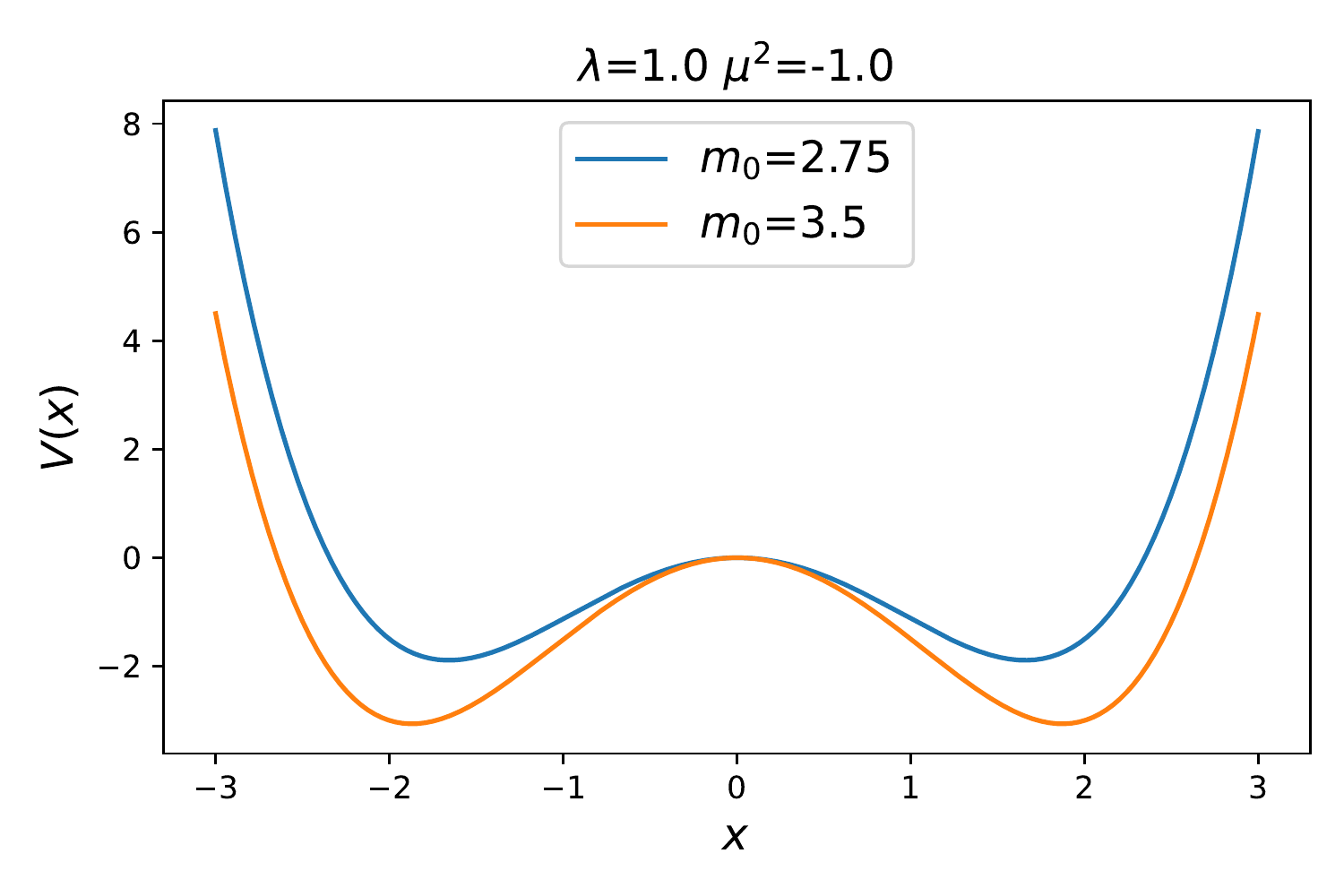}
    \includegraphics[width=0.48 \textwidth]{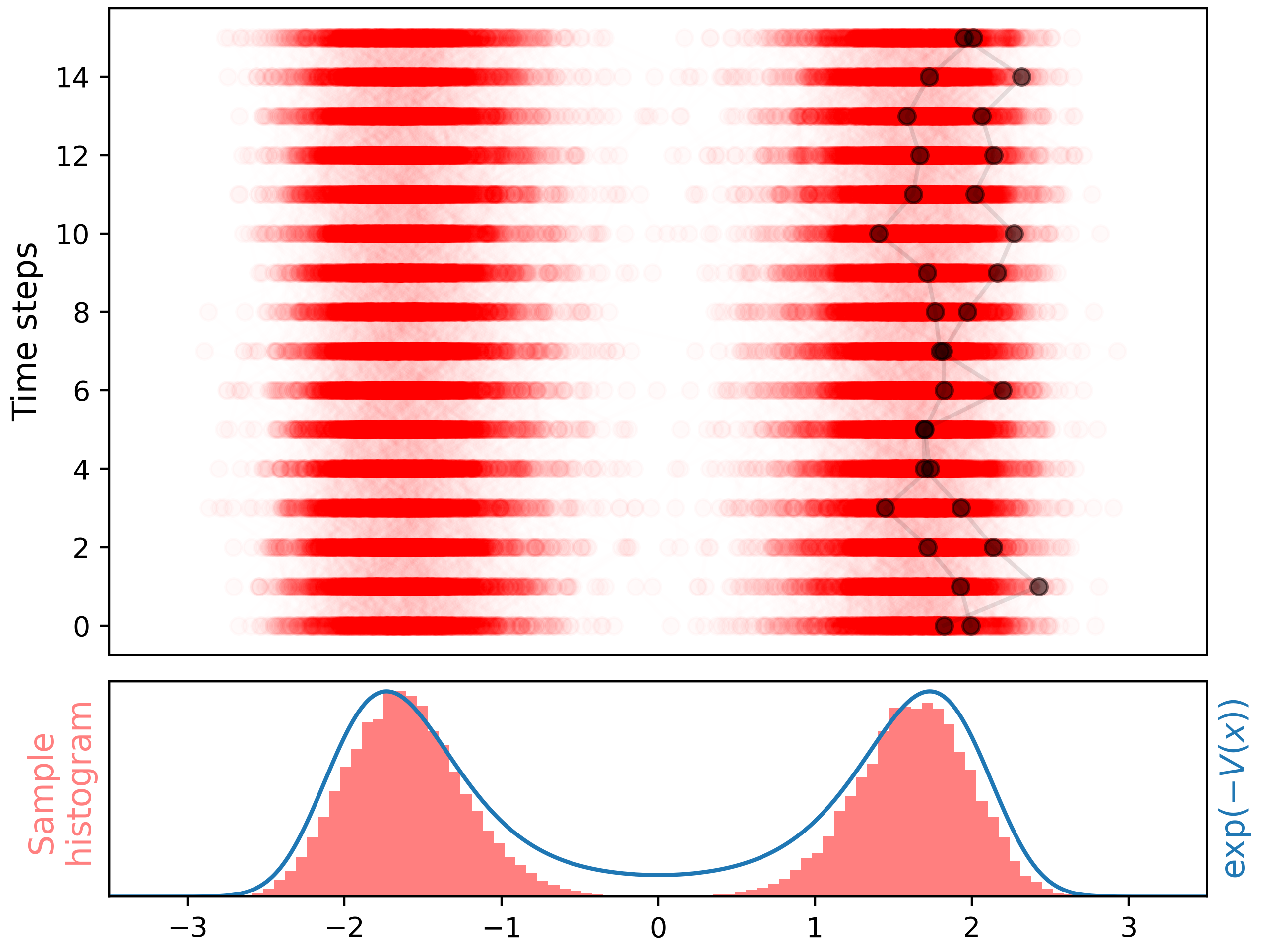}
    \caption{\textbf{Left}: Double-well potential for various choices of the mass $m_0$. \textbf{Right}: Visualization of samples from a trained model.}
    \label{fig:dw-samples-after-training}
\end{figure}

In quantum mechanics, the position of the particle at euclidean time $t$, and thus its path $x(t)$, is a random variable. For a discretized path $x = (x_0, x_1, ..., x_{T-1})$, the corresponding density is given by
\begin{align}
    p(x) = \frac{1}{Z} \exp(-S(x))
\end{align}
where $Z = \int d^T x \exp(-S(x))$ is the partition function and the action is given by
\begin{equation}
	S(x) = a \sum_{t=0}^{T-1} \left( \frac{ m_0}{2} \left( x_{t+1} - x_{t} \right)^2 + V(x_t)  \right)
\end{equation}
with periodic boundary conditions for $x_i$ and $a$ denoting the lattice spacing. The double-well potential $V$ is defined by
\begin{equation}
	V(x) = 	\frac{ m_0 \mu^2}{2} x^2 + \frac{\lambda}{4} x^4 \label{eq:dw-pot} \,,
\end{equation}
where the mass parameters $m_0$ and $\mu^2$, as well as the coupling $\lambda$ control the shape of the potential, see the left part of Figure~\ref{fig:dw-samples-after-training}. 

\subsection{Forward and Reverse Effective Sampling Size}\label{sec:ess}
In our numerical experiments, we evaluate the degree to which the model $q_\theta$ approximates the target density $p$. As explained in Section~\ref{sec:samplingwithflows}, the effective sampling size \eqref{eq:ess} is a natural metric to quantify this. 

We can estimate the effective sampling size using two approaches \cite{hackett2021flow, nicoli2021machine}:
\begin{itemize}
    \item Reverse estimation uses samples from the flow
    \begin{align*}
        \textrm{ESS} = \frac{1}{\mathbb{E}_q \left[ {w}^2 \right]} \approx \frac{1}{\frac{1}{N} \sum_{i=1}^N \hat w(x_i)^2} \,, && x_i \sim q_\theta \,.
    \end{align*} 
    \item Forward estimation uses samples from the target density $p$ 
    \begin{align*}
        \textrm{ESS} = \frac{1}{\mathbb{E}_q \left[ {w}^2 \right]} = \frac{1}{\mathbb{E}_p \left[ {w} \right]} \approx \frac{1}{\frac{1}{N} \sum_{i=1}^N \hat w(x_i)} \,, && x_i \sim p \,.
    \end{align*}
\end{itemize}
As discussed in Section~\ref{sec:forwardTraining}, avoiding mode-collapse is of critical importance for neural importance sampling. However, the reverse estimator of the effective sampling size is not sensitive to mode-collapse as it just uses samples from the model. This is different for the forward estimator which in turn however has the disadvantage that it requires samples from the target density $p$ which may be very costly to generate. It can therefore be challenging to detect mode-collapse - especially in situations for which MCMC methods fail. 

For the particle in the double-well potential, we can use an overrelaxed Hybrid-Monte-Carlo algorithm to generate ground truth samples of the target $p$. These samples are then used to estimate the forward effective sampling size. This allows us to detect whether a certain training procedure leads to mode-collapse and thus quantify the degree of approximation correctly. The details of running the MCMC can be found in \Secref{app:exp-details}.

\subsection{Discussion of Results}
\paragraph{Setup:} we train a flow with RealNVP couplings for lattices of size $L \in \{8, 16, 32, 64\}$ for each value of the mass $m_0\in \{2.75, 3.25\}$. We fix the other parameters to $\lambda=1$ and $\mu^2=-1$. For forward and reverse KL training, we use ReinfPQ \eqref{eq:forwardReinforce} and the RepQP estimator \eqref{eq:reparamRev} as the baseline respectively because these are the most widely used loss functions. We ensure that the baseline and path-gradient estimators use the same walltime for training in order to ensure fair comparison and repeat training five times for uncertainty estimation. We then estimate the forward effective sampling size as described in the previous section. For a detailed description of the architecture and training procedure, we refer to the Appendix \ref{app:exp-details}.  

\begin{figure}
    \centering
    \includegraphics[width=0.48 \textwidth]{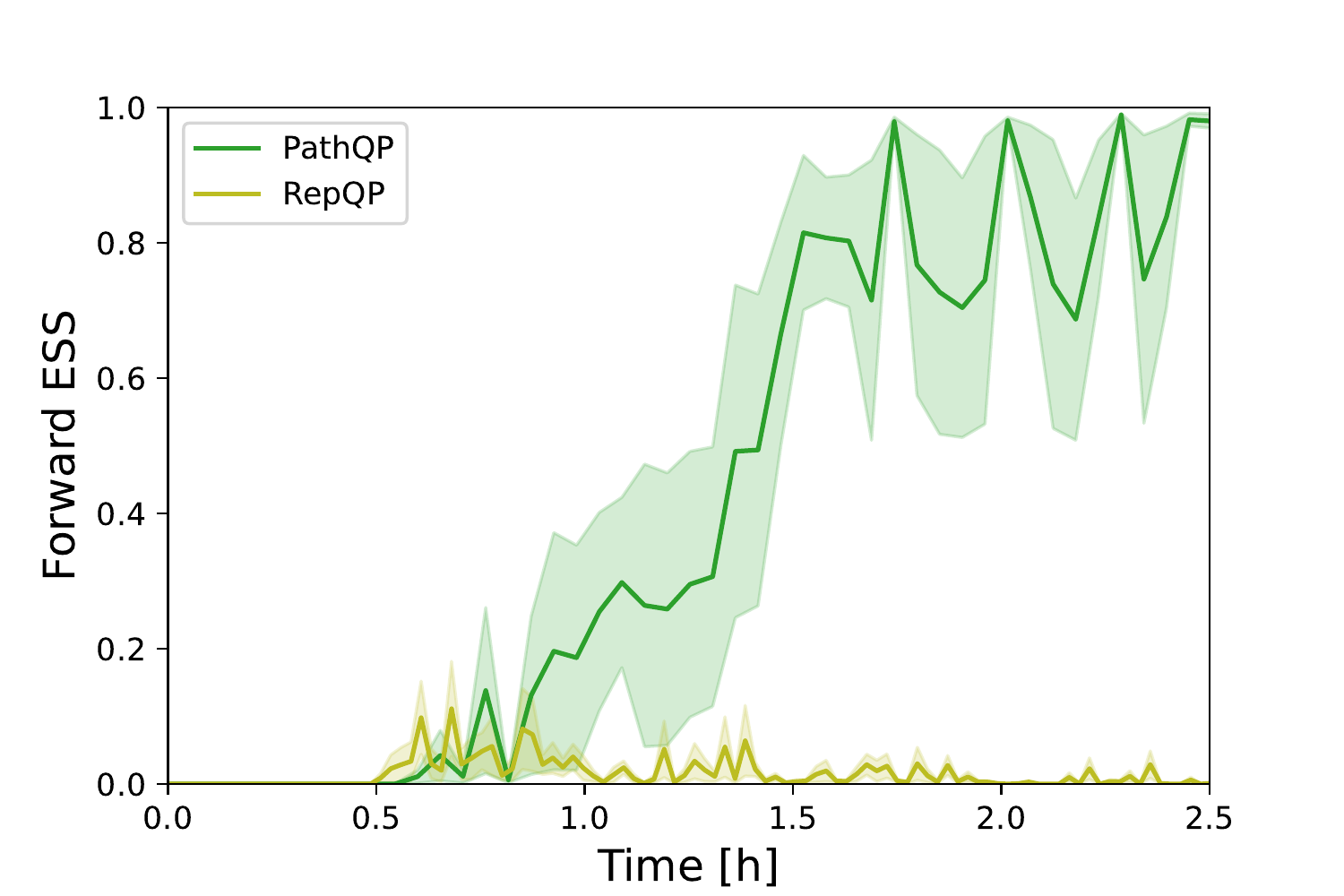}
    \includegraphics[width=0.48 \textwidth]{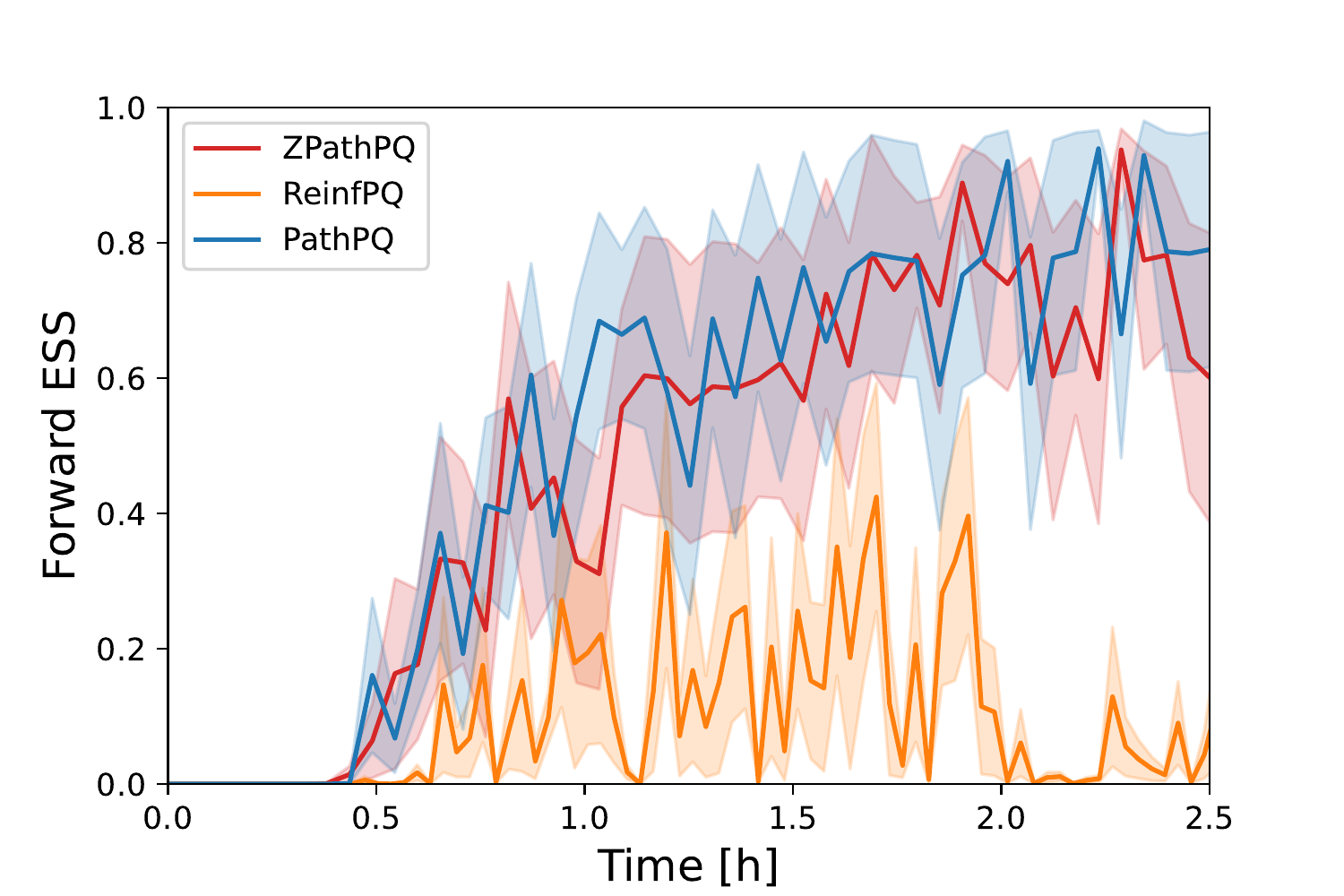}
    \caption{\textbf{Left}: Training with the reverse KL estimators for $m_0=2.75$. \textbf{Right}: Training with the forward KL estimators for $m_0=3.0$. The path gradient estimators clearly outperform their baselines. Bold line denotes the mean and shaded area is the standard error over five training runs.}
    \label{fig:ess_over_training}
\end{figure}

\begin{figure}
    \centering
    \includegraphics[width=0.48 \textwidth]{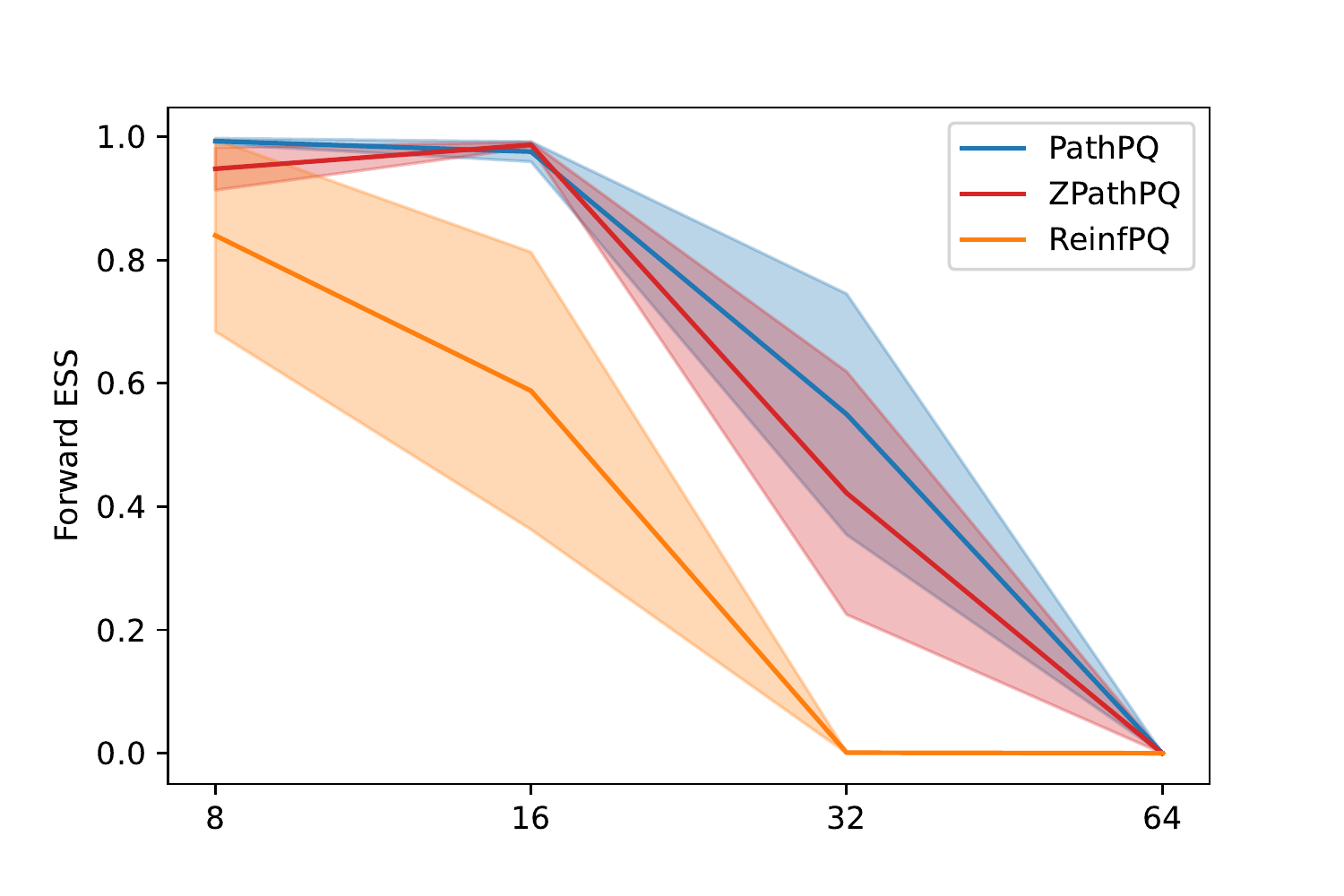}
    \includegraphics[width=0.48 \textwidth]{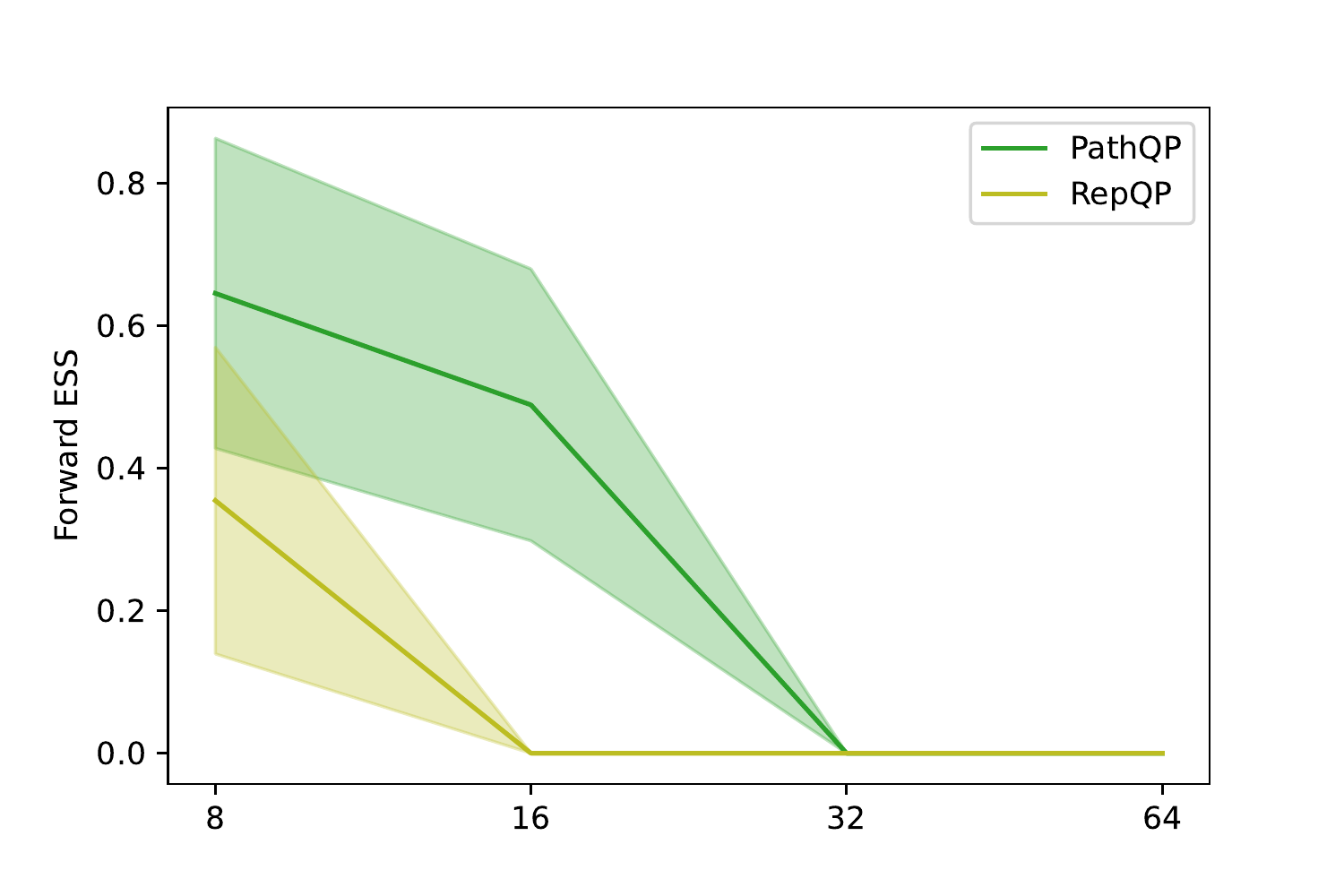}
    \caption{Forward ESS for $m_0=3.0$. \textbf{Left}: Forward KL training using both path-gradient estimators \eqref{eq:vanilla} and \eqref{eq:dreg} compared to the reinforce-based baseline \eqref{eq:reinforce}. \textbf{Right}: Reverse KL-training using the path-gradient estimator \eqref{eq:pathqp} compared to the base of the reparameterized estimator \eqref{eq:reparamRev}. The path-gradient-based estimators consistently outperform their baselines.}
    \label{fig:FW-Ess-vs_dims}
\end{figure}

\paragraph{Approximation quality:} Figure~\ref{fig:ess_over_training} shows an example of training as a function of walltime. The path gradient estimators clearly outperform the standard ones for both forward and reverse KL divergence. Figure~\ref{fig:FW-Ess-vs_dims} compares the path-gradient estimators to a suitable baseline for other choices of the mass $m_0$ and demonstrates that the superior performance of the path gradient estimators is not due to a particular choice of the mass $m_0$. Our experiments confirm the observation of related work \citep{dhaka2021challenges, geffner2020difficulty}, that optimizing the forward KL only works up to moderately high dimensions. Nevertheless using the path-wise gradients increases the feasible number of dimensions for which optimizing the forward KL is still a viable option.

\paragraph{Mode-collapse:} in order to analyze mode-collapse, we compare the forward and reverse estimate of the effective sampling sizes for path-gradient training and its baseline. For Table~\ref{tab:dw-res-table-3.0}, we see that reverse training suffer from mode-collapse starting from relatively small lattice sizes in contrast to forward training as shown by the discrepancy of forward and reverse estimate of the effective sampling size. This underscores the superiority of forward training as it is crucial that mode-collapse is avoided for the statistical consistency of neural importance sampling. 
Furthermore, both for forward and reverse training, the path-gradient-based methods suffer substantially less from mode-collapse. In particular, we observe that the $Z$PathPQ estimator \eqref{eq:dreg} seems to be less susceptible to mode-collapse. As we increase the mass $m_0$, and therefore move deeper into the broken phase, the mode-collapse becomes more severe across all methods, see Table~\ref{tab:dw-res-table-3.25}. Nevertheless, the path-gradient-based methods allow us to estimate at parameter values for which the standard methods fail.
\paragraph{Training Phases:} Figure~\ref{fig:DReG-Initial-Phase} demonstrates that the $Z$PathPQ estimator \eqref{eq:dreg} can suffer from vanishing gradients in the initial phase of training. This indeed confirms our theoretical analysis in Section~\ref{sec:theorytrainingphases}. It may therefore be advisable to start training with the the PathPQ estimator \eqref{eq:forwardPG}. From Figure~\ref{fig:emp-STL-gradnorm}, it can be seen that the norm of the PathQP estimator \eqref{eq:pathqp} indeed vanishes in the final phase of training while the standard RepQP estimator \eqref{eq:reparamRev} is non-vanishing. This is to be expected since the latter contains the score term \eqref{eq:scoreTerm} which only vanishes in expectation.  

\begin{table}
\centering
\begin{tabular}{l|r||lll|ll}
	{ESS} & {$d$} & {ReinfPQ} & {ZPathPQ} & {PathPQ} & {RepQP} & {PathQP}  \\
	\hline
\multirow[c]{5}{*}{FW ESS} 
 & 8 & \textbf{0.84} $\pm$ .14 & \textbf{0.95} $\pm$ .03 & \textbf{0.99} $\pm$ .00 & 0.03 $\pm$ .02 & \textbf{0.65} $\pm$ .19 \\
 & 16 & \textbf{0.59} $\pm$ .20 & \textbf{0.99} $\pm$ .00 & \textbf{0.98} $\pm$ .01 & 0.00 $\pm$ .00 &0.49 $\pm$ .17 \\
 & 32 & 0.00 $\pm$ .00 & \textbf{0.42} $\pm$ .18 & \textbf{0.55} $\pm$ .17 & {0.00} $\pm$ .00 & {0.00} $\pm$ .00 \\
 & 64 & \textbf{0.00} $\pm$ .00 & \textbf{0.00} $\pm$ .00 & \textbf{0.00} $\pm$ .00 & \textbf{0.00} $\pm$ .00 & \textbf{0.00} $\pm$ .00 \\
 \hline
\multirow[c]{5}{*}{Rev ESS} 
 & 8 & 1.00 $\pm$ .00 & \textbf{1.00} $\pm$ .00 & \textbf{1.00} $\pm$ .00 & 0.99 $\pm$ .00 & \textbf{1.00} $\pm$ .00 \\
 & 16 & 0.99 $\pm$ .00 & \textbf{0.99} $\pm$ .00 & \textbf{1.00} $\pm$ .00 & 0.06 $\pm$ .04 & 0.99 $\pm$ .00 \\
 & 32 & \textbf{0.96} $\pm$ .00 & \textbf{0.98} $\pm$ .00 & \textbf{0.98} $\pm$ .00 & 0.78 $\pm$ .13 & 0.80 $\pm$ .07 \\
 & 64 & 0.93 $\pm$ .01 & 0.95 $\pm$ .00 & 0.92 $\pm$ .03 & 0.90 $\pm$ .04 & \textbf{0.97} $\pm$ .00 \\
\end{tabular}
\caption{Results of training a RealNVP for $m_0=3$. Note that only the forward estimators can detect mode-collapse as discussed in Section~\ref{sec:ess}. As a result, a large value of the reverse with a corresponding small value for the forward ESS is a clear indication of mode-collapse. The path-gradient estimators therefore not only consistently outperform the baselines but also are significantly more robust to mode-collapse. Statistical significance was tested with the Wilcoxon signed-rank test.}
    \label{tab:dw-res-table-3.0}
\end{table}

\begin{table}
\centering
 \begin{tabular}{l|r||lll|ll}
 	{ESS} & {$d$} & {ReinfPQ} & {ZPathPQ} & {PathPQ} & {RepQP} & {PathQP}  \\
 	\hline
\multirow[c]{5}{*}{FW ESS} 
 & 8 & 0.40 $\pm$ .22 &\textbf{0.81} $\pm$ .16 & \textbf{1.00 }$\pm$ .00 & 0.00 $\pm$ .00 & 0.18 $\pm$ .16 \\
 & 16 & 0.00 $\pm$ .00 & \textbf{0.86} $\pm$ .10 & \textbf{0.79} $\pm$ .18 & 0.00 $\pm$ .00 & 0.00 $\pm$ .00 \\
 & 32 & 0.00 $\pm$ .00 &\textbf{0.14} $\pm$ .13 & 0.00 $\pm$ .00 & 0.00 $\pm$ .00 & 0.00 $\pm$ .00 \\
 & 64 & \textbf{0.00} $\pm$ .00 & \textbf{0.00} $\pm$ .00 & \textbf{0.00} $\pm$ .00 & \textbf{0.00} $\pm$ .00 & \textbf{0.00} $\pm$ .00 \\
 \hline
\multirow[c]{5}{*}{Rev ESS} 
 & 8 & 0.99 $\pm$ .00 & \textbf{1.00} $\pm$ .00 & \textbf{1.00} $\pm$ .00 & \textbf{0.77} $\pm$ .17 & 0.93 $\pm$ .06 \\
 & 16 & 0.99 $\pm$ .00 & \textbf{0.99} $\pm$ .00 & \textbf{0.99} $\pm$ .00 & \textbf{0.99} $\pm$ .00 & \textbf{0.60} $\pm$ .16 \\
 & 32 & \textbf{0.92} $\pm$ .03 & \textbf{0.96} $\pm$ .01 & \textbf{0.97} $\pm$ .01 & \textbf{0.98} $\pm$ .00 & \textbf{0.91} $\pm$ .06 \\
 & 64 & 0.94 $\pm$ .00 & 0.95 $\pm$ .00 & 0.96 $\pm$ .00 & 0.87 $\pm$ .06 & \textbf{0.98} $\pm$ .00 \\
\end{tabular}
    \caption{Same as Table~\ref{tab:dw-res-table-3.0} but for a relatively large mass of $m_0=3.25$. At this point in parameter space, the modes of the distribution are separated by a pronounced action barrier which can lead to mode collapse.}
    \label{tab:dw-res-table-3.25}
\end{table}

\begin{figure}[ht!]
    \centering
    \includegraphics[width=0.95\textwidth]{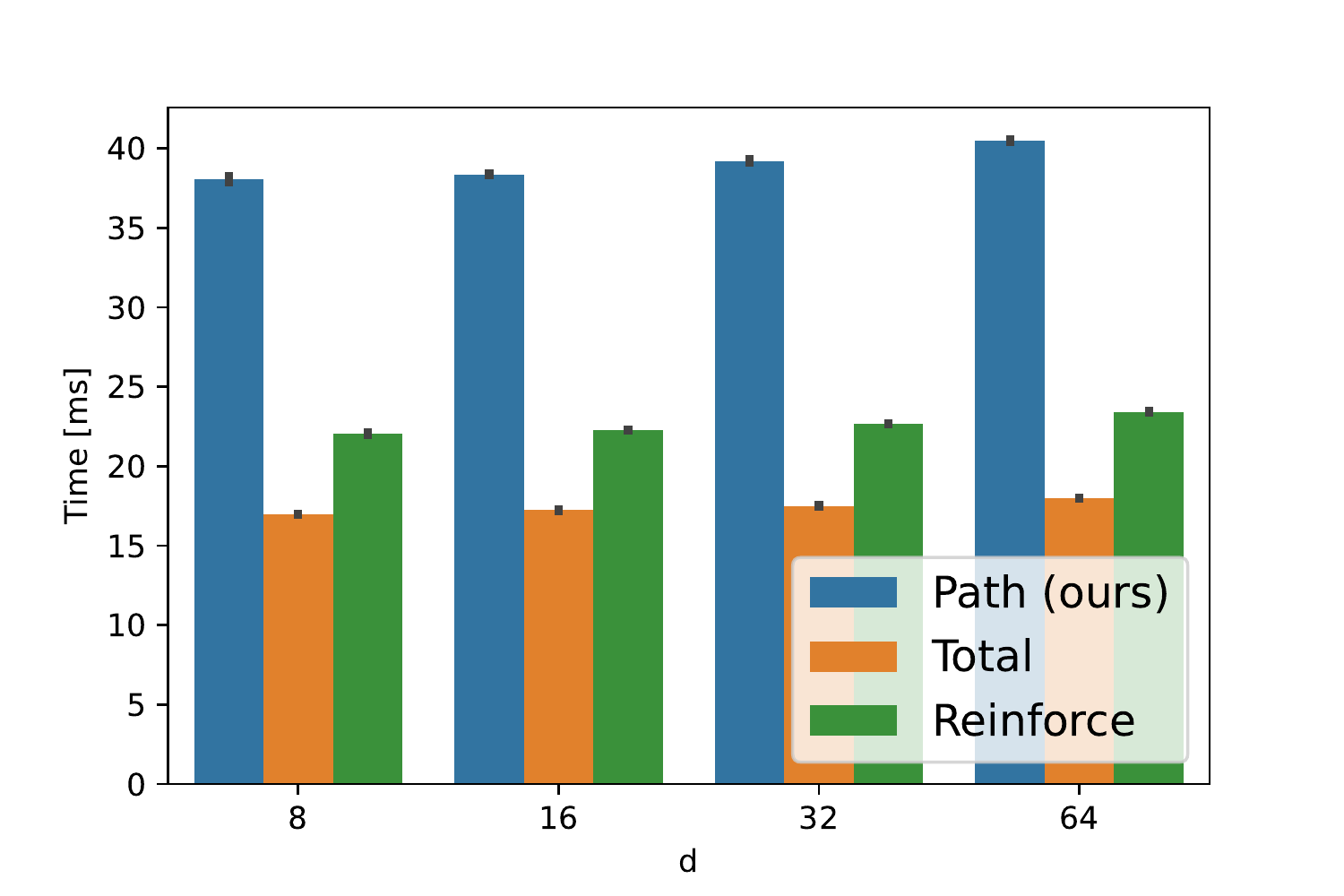}
    \caption{Runtime per iteration for calculation of the path, total, and reinforce gradient. The path-gradient has roughly twice the computational cost at the same memory requirements. This overhead is more than compensated by faster overall training convergence.}
    \label{fig:runtime}
\end{figure}

\paragraph{Runtime:} as discussed in Section~\ref{sec:implementation}, our estimator for the path gradient has the same memory requirements as the standard estimator for the total derivative. This is crucial as it allows us to train with the same batch-size. However, we expect double runtime per iteration. This is indeed confirmed by our numerical experiments, see Figure~\ref{fig:runtime}. In order to account for this, the training using the PathQP and ReinfPQ baselines are therefore run for twice the number of iterations. This ensures a fair comparison as all estimators have the same overall training time.   

\section{Conclusion}
Path-gradient estimators bring substantial improvements for training normalizing flows in the context of neural importance sampling. In this work, we have proposed an algorithm to calculate path gradient estimators that can be applied as a drop-in replacement of standard estimators for any invertible normalizing flow.
Crucially, our algorithm has the same memory requirements as standard approaches and thus allows us to to use the same batch size for training, which is essential for training with MC methods.
Furthermore, the lower variance of the path-wise gradient estimators not only leads to faster convergence during training, but also better overall approximation quality. 

The path-wise gradient estimators allow us to apply the flows to higher-dimensional problems by pushing the limit up to which the inclusive forward KL is applicable. Our experiments have demonstrated the favorable behavior of the forward KL with respect to mode collapse in moderately high dimension, which enables us to tackle mode collapse without any prior domain knowledge of the problem at hand.
We have analyzed our estimators theoretically and shown that they have lower variance in the limit of perfect approximation. We furthermore theoretically compared the properties of the forward estimators in the initial and final phases of training. 
We expect that the estimator proposed in this work will become the new de-facto standard for training normalizing flows on variational inference tasks due to its superior performance and implementation simplicity.

For future work, it would be interesting to theoretically prove the lower variance of the path-gradient estimators off the limit of perfect approximation as is strongly suggested by our experiments. Furthermore, it would be very desirable to construct an estimator which is as fast as the one derived in \cite{vaitl2022path} since their proposal is more performant than ours but unfortunately completely limited to the special case of a continuous normalizing flow. 

\subsubsection*{Acknowledgments}
K.A.N., S.N. and P.K. are supported by the German Ministry for Education and Research (BMBF) as BIFOLD - Berlin Institute for the Foundations of Learning and Data under grants 01IS18025A and  01IS18037A.

\bibliography{iclr2021_conference}

\appendix

\section{Variance of the Reinforce Estimator} \label{app:reinf}

The ReinfPQ estimator was defined in \eqref{eq:forwardReinforce} as
\begin{align}
     \frac{d}{d \theta} \textrm{KL} (p, q_\theta)
     &\approx \frac{1}{N} \sum_{i=1}^N \frac{\tilde{w}_i}{\hat{Z}} \frac{\partial}{\partial \theta} \log \left( \tilde{w}_i \right) \\
     &= \frac{1}{N} \sum_{i=1}^N \frac{e^{-S(x_i)}}{ q_\theta (x_i) \hat{Z}} \frac{\partial}{\partial \theta} \log q_\theta(x_i) .
     \notag
\end{align}
Its second moment is therefore given by
\begin{align*}
    \frac{1}{N} \,  \mathbb{E}_{x \sim q_\theta} \left[ {\hat{w}(x)}^2 \left(\frac{\partial}{\partial \theta} \log q_\theta(x) \frac{\partial}{\partial \theta} \log q_\theta(x)\right) \right] \,.
\end{align*}
In the limit of perfect approximation, i.e., $q_\theta (x) = p(x)$ for any $x \in \mathcal{X}$, it holds that ${\hat{w}(x)} = 1$ and thus the covariance of the ReinfPQ estimator converges to 
the Fisher information of the variational distribution
\begin{align}
    \frac{1}{N} \,  \underbrace{\mathbb{E}_{x \sim q_\theta} \left[ \frac{\partial}{\partial \theta} \log q_\theta(x) \frac{\partial}{\partial \theta} \log q_\theta(x) \right]}_{=\mathcal{I}(\theta)} =  \frac{1}{N}  \mathcal{I}(\theta) \,,
\end{align}
and thus is generically non-vanishing even in the limit of perfect approximation.

\section{Asymptotic Behavior of Path Gradient Estimators of Forward KL Divergence}
\label{app:Asympt-behaviour}

Here, we analyze the bias and variance of the PathPQ \eqref{eq:vanilla} and ZPathPQ \eqref{eq:dreg} estimators, which are defined as
\begin{align}
\text{PathPQ}_N 
&= - \frac 1 N \sum_{i=1}^N 
\frac {\tilde w_i}{\frac 1 N \sum_{j=1}^N \tilde w_j }
\blacktriangledown_\theta  \log \tilde w_i  \notag, \\
\text{ZPathPQ}_N 
&= \text{PathPQ}_N + \frac 1 {N^2} \sum_{i=1}^N  
\left(\frac {\tilde w_i}{ \frac 1 N \sum_{j=1}^N \tilde w_i } \right)^2
\blacktriangledown_\theta \log  \tilde w_i  .
\notag
\end{align}
Throughout this appendix,
we use
a shorthand notation for $\tilde{w}_i = \tilde{w}(x_i) = \tilde{w}(g_\theta(z_i))$.
Using $w_i = \tilde w_i / Z = p(x_i) / q_{\theta}(x_i)$, we can rewrite the estimators above as
\begin{align}
\text{PathPQ}_N 
&= - \frac 1 N \sum_{i=1}^N 
\frac {1}{\frac 1 N \sum_{j=1}^N  w_j }
\blacktriangledown_\theta   w_i  
\label{eq:PATHPQEstimatorApp}, \\
\text{ZPathPQ}_N 
&= \text{PathPQ}_N + \frac 1 {N^2} \sum_{i=1}^N  
\frac { w_i}{\left( \frac 1 N \sum_{j=1}^N  w_i\right)^2 } 
\blacktriangledown_\theta    w_i ,
\label{eq:ZPATHPQEstimatorApp}
\end{align}
where we used $  w_i\blacktriangledown_\theta  \log  w_i = \blacktriangledown_\theta   w_i $.
Note that $\E_{q_{\theta}}[w] = 1$.

\subsection{Bias}

Let ${\epsilon \equiv \frac 1 N \sum_{i=1}^N (1 - w_i)}$.  Then, $\E_{q_{\theta}}[\epsilon] = 0$ and $\epsilon = \mathcal O_p(N^{-1/2})$, and therefore
\begin{align}
 \frac{1}{\frac 1 N \sum_{j=1}^N    w_j} 
										&= \left( 1 - \epsilon\right)^{-1}   = 1 + \epsilon + \epsilon^2 +\epsilon^3 + \mathcal O_p(N^{-2}) .
										\label{eq:EpsilonDecomposition}
\end{align}
Since $\{w_i\}_{i=1}^N$ are independent,
the following hold for any function $\kappa(\cdot)$:
\begin{align}
    \E_{q_{\theta}}[\kappa( w_i) \epsilon] &= \E_{q_{\theta}}\left[\kappa (w_i) \frac 1 N \sum_{j=1}^N (1 - w_j)\right] \notag\\
    &= \frac 1 N  \E_{q_{\theta}}[\kappa (w_i) (1 - w_i)], \label{eq:EpsilonKappaOne}\\
\E_{q_{\theta}}[\kappa( w_i) \epsilon^2] &= \E_{q_{\theta}} \left[\kappa( w_i) \frac 1 {N^2} \left(\sum_{j=1}^N \sum_{k=1}^N (1 - w_j) (1 - w_k ) \right) \right] \notag\\
&= \E_{q_{\theta}} \left[\kappa (w_i) \frac 1 {N^2} \left((1 - w_i )^2 + \sum_{j\neq i} (1 - w_j )^2 + \sum_{j=1}^N \sum_{k \neq j} (1 - w_j) (1 - w_k )\right) \right] \notag\\
&= \E_{q_{\theta}} \left[\kappa (w_i)\frac 1 {N^2} \left((1 - w_i )^2 + (N-1) \E_{w \sim q_{\theta}}[(1 - w)^2] \right)\right] \notag\\
&=  \frac{1}{N} \E_{w \sim q_{\theta}} \left[\kappa (w_i) \right]
 \E_{w \sim q_{\theta}}[(1 - w)^2 ] 
+ \E_{q_{\theta}} \left[\kappa (w_i) \cdot \mathcal O_p(N^{-2}) \right]
, \label{eq:EpsilonKappaTwo}\\
\E_{q_{\theta}}[\kappa (w_i) \epsilon^3] &= \E_{q_{\theta}} \left[\kappa (w_i) \frac 1 {N^3} \left((1 - w_i)^3 + \sum_{j\neq i} ((1 - w_j)^3 + 3 (1 - w_i )(1 - w_j )^2 )\right) \right] \notag\\
&= \E_{q_{\theta}} \left[\kappa (w_i) \frac 1 {N^3} \left((1 - w_i )^3 + (N-1) \E_{w \sim q_{\theta}}\left[ ((1 - w )^3 + 3 (1 - w_i )(1 - w )^2 )\right]\right) \right] \notag\\
&= \E_{ q_{\theta}} \left[\kappa (w_i) \cdot \mathcal O_p(N^{-2}) \right].
\label{eq:EpsilonKappaThree}
\end{align}

\paragraph{PathPQ}
By using  Eqs.\eqref{eq:EpsilonDecomposition}-\eqref{eq:EpsilonKappaThree}, 
the bias of the PathPQ estimator \eqref{eq:PATHPQEstimatorApp} 
from the true gradient $-	\E_{w \sim q_{\theta}}[\blacktriangledown_\theta   w]$
is evaluated as
\begin{align}
	\E_{q_{\theta}} & [\text{PathPQ}_N] 
	+ 	\E_{w \sim q_{\theta}}[\blacktriangledown_\theta   w] 
\notag\\
	&=   - \frac 1 N \left( \E_{w \sim q_{\theta}}[ (1 - w) \blacktriangledown_\theta w]
+   \E_{w \sim q_{\theta}}\left[(1 - w)^2 \right] \E_{q_{\theta}}[\blacktriangledown_\theta w] \right)
	+ \E_{w \sim q_{\theta}}\left[  \mathcal O_p(N^{-2}) \cdot \blacktriangledown_\theta w \right]\, . 
	\label{eq:biasPathPQ}
\end{align}
The bias of the PathPQ estimator is of $N^{-1}$ times smaller scale than the true gradient.  However, the leading term vanishes in the converging phase where $q_{\theta} (x) \approx p(x)$ holds.  For example, if $w = 1 + \mathcal O_p (N^{-1})$, the bias is of $N^{-2}$ times smaller scale.
We suppose that this, however, does not affect the practical training performance, because the estimation error is dominated by the variance, as will be shown in Appendix~\ref{sec:A.Asymptotic.Variance}.


\paragraph{ZPathPQ}

Similarly, by using  Eqs.\eqref{eq:EpsilonDecomposition}-\eqref{eq:EpsilonKappaThree},
the expectation value of the second term of the ZPathPQ estimator \eqref{eq:ZPATHPQEstimatorApp}
is evaluated as
\begin{align}
\E_{q_{\theta}}\left[
 \frac 1 {N^2} \sum_{i=1}^N  
\frac { w_i}{\left( \frac 1 N \sum_{j=1}^N  w_i\right)^2 } 
\blacktriangledown_\theta    w_i \right]
	&=  \frac 1 {N} \E_{w \sim q_{\theta}}[w \blacktriangledown_\theta w] + \E_{w \sim q_{\theta}}\left[  \mathcal O_p(N^{-2}) \cdot \blacktriangledown_\theta w   \right].
	\notag
\end{align}
Therefore the bias of the ZPathPQ estimator is given as
\begin{align}
	\E_{q_{\theta}} & [\text{ZPathPQ}_N] 
	+ 	\E_{w \sim q_{\theta}}[\blacktriangledown_\theta   w] 
\notag\\
	&=   - \frac 1 N \left( \E_{w \sim q_{\theta}}[ (1 - 2w) \blacktriangledown_\theta w]
+   \E_{w \sim q_{\theta}}\left[(1 - w)^2 \right] \E_{q_{\theta}}[\blacktriangledown_\theta w] \right)
	+ \E_{w \sim q_{\theta}}\left[ \mathcal O(N^{-2}) \cdot \blacktriangledown_\theta w    \right]\, . 
	\label{eq:biasZPathPQ}
\end{align}
A notable difference from the PathPQ estimator is that the leading order term does not vanish when $w \approx 1$, and therefore the bias is always $N^{-1}$ times smaller than the true gradient.

\subsection{Variance}
\label{sec:A.Asymptotic.Variance}
We analyze the asymptotic behavior of the variance of gradient estimators
by using the Delta Method (see e.g. Theorem 5.5.28 in  \citet{casella2021statistical} or  Paragraph 4.9 in \citet{small2010expansions}):
\begin{theorem} \citep{small2010expansions}
Assume that $f(x,y, z)$ is a differentiable function, 
and $\{(X_i, Y_i, Z_i)\}_{i=1}^N$ are independently and identically distributed.
Assume that the distribution of $(X_i, Y_i, Z_i)$ satisfies the conditions for the central limit theorem, and hence 
 $\overline X = \frac 1 N \sum_{i=1}^N X_i$, $\overline Y = \frac 1 N \sum_{i=1}^N Y_i$, and $\overline Z = \frac 1 N \sum_{i=1}^N Z_i$ are normally distributed in the asymptotic limit.
Then it holds that
\begin{align}
\var(f(\overline X,\overline Y,\overline Z)) 
& = f_x^2 \var(\overline X) 
+ f_y^2 \var(\overline Y) 
+ f_z^2 \var(\overline Z) 
\notag\\
& \qquad
+ 2 f_x f_y \Cov(\overline X,\overline Y) 
+ 2 f_y f_z \Cov(\overline Y,\overline Z) 
+ 2 f_z f_x \Cov(\overline Z,\overline X) 
\notag\\
& \qquad
+ \mathcal O(N^{-2}) \, , \label{eq:Delta2vars}
\end{align}
where 
$$
(f_x, f_y, f_z) = \left(	\frac{\partial f}{\partial x}, 	\frac{\partial f}{\partial y}, 	\frac{\partial f}{\partial z}\right) \Big|_{(x,y, z) = (\E[\overline X], \E[\overline Y], \E[\overline Z])}
$$
are
the derivatives evaluated at $(x,y, z) = (\E[\overline X], \E[\overline Y], \E[\overline Z])$.
\end{theorem}

\paragraph{PathPQ}
The PathPQ estimator \eqref{eq:PATHPQEstimatorApp} is a ratio estimator with two variables as
$$
\text{PathPQ}_N \equiv
f(\overline X,\overline Y) = -  \underbrace{ \frac 1 {\frac 1 N \sum_{j=1}^N w_j}}_{1/\overline Y} \underbrace{\frac 1 N \sum_{i=1}^N 	\blacktriangledown_\theta  w_i}_{\overline X}  = - \frac {\overline X} {\overline Y}.
$$
Since
\begin{align*}
	f_x&= - \frac{1}{\E[\overline Y]} =- \frac 1 {\E_{q_{\theta}}[w]} = -1,\\
	f_y &=  \frac {\E[\overline X]} {\E[\overline Y^2]} =  \frac {\E_{q_{\theta}}[\blacktriangledown_\theta w]}{\E_{q_{\theta}}[w]^2}, \\
	\Var[\overline X] &= \frac 1 N \Var_{q_{\theta}}[\blacktriangledown_\theta w], \\
	\Var[\overline Y] &= \frac 1 N \Var_{q_{\theta}}[w], \\
		\Cov[\overline X,\overline Y] 
			  &=\frac 1 N \Cov_{q_{\theta}}[ \blacktriangledown_\theta w, w], 
\end{align*}
Eq.\eqref{eq:Delta2vars} gives
\begin{align}
			 \Var(\text{PathPQ}_N) &= \Var(f(\overline X,\overline Y)) \notag \\
			 &= \frac 1 {\E[\overline Y]^2} \Var(\overline X) 
			 + \frac {\E[\overline X]^2}{\E[\overline Y]^4} \Var[\overline Y]
			 - 2 \frac 1 {\E[\overline Y]} \frac {\E[\overline X]} {\E[\overline Y]^2} \cov[\overline X,\overline Y]		 
			 +  O(N^{-2})\notag  \\				 
		      &= \frac{1}{N} \Var_{q_{\theta}}[\blacktriangledown_\theta w] 
		      		      +
		      \frac{\E_{q_{\theta}}[\blacktriangledown_\theta w]^2}{N } \Var_{q_{\theta}}[w]
 \notag \\
		      & \hspace{1em}
		      - 2 \frac {\E_{q_{\theta}}[\blacktriangledown_\theta w]} {N } \cov_{q_{\theta}}[\blacktriangledown_\theta w, w]		      
 +  \mathcal O (N^{-2}).
  \label{eq:VariancePathPQ}
\end{align}
\paragraph{ZPathPQ}
ZPathPQ \eqref{eq:PATHPQEstimatorApp} has an additional term to PathPQ, i.e.,
\begin{align*}
\text{ZPathPQ}_N \equiv
	f'(\overline X,\overline Y, \overline {Z}) &= f(\overline X,\overline Y) + g(\overline  Y, \overline{ Z}), \qquad \mbox{ where} \\
	g(\overline Y, \overline{ Z}) &=  \underbrace{\frac{1}{\left(\frac 1 N \sum_{j=1}^k  \sample w j\right)^2}}_{\overline{ Y}^{-2}} \underbrace{\frac 1 {N^2} \sum_{i=1}^k \sample w i \blacktriangledown_\theta \sample w i}_{\overline{ Z}} = \frac{\overline {Z}}{\overline{ Y}^2}. 
\end{align*}
Since 
\begin{align*}
f'_x &= f_x\, , \\
f'_{y} &= f_y + 2 \frac{\E[\overline {Z}]}{\E[\overline{ Y}^3]} =  \frac {\E_{q_{\theta}}[\blacktriangledown_\theta w]}{\E_{q_{\theta}}[w]^2} +  \frac 2 N \frac{\E_{q_{\theta}}[w \blacktriangledown_\theta w]}{E_{q_{\theta}}[w]^3}\, , \\
f'_{z} &=- \frac{1}{\E[{\overline Y}]^2} = -\frac{1}{\E_{q_{\theta}}[{w}]^2}\, , \\
\Var(\overline {Z}) &= \frac 1 {N^3} \Var_{q_{\theta}}(w \blacktriangledown_\theta w) \, ,\\
\Cov(\overline {Y}, \overline {Z}) &= \frac 1 {N^2} \Cov_{q_{\theta}}[w, w \blacktriangledown_\theta w],\\
\Cov(\overline {Z}, \overline {X}) &= \frac 1 {N^2} \Cov_{q_{\theta}}(w \blacktriangledown_\theta w, \blacktriangledown_\theta w)\, ,
\end{align*}
Eq.\eqref{eq:Delta2vars} gives
\begin{align}
\Var(\text{ZPathPQ}_N) &= \Var(f'(\overline X, \overline Y, \overline {Z})) \notag \\
 &= \var(\overline X) f_X'^2 + \var(\overline Y) f_Y'^2 + \var(\overline {Z}) f_{{z}}'^2 
 \notag \\
		     & \hspace{1em} 
		     + 2 f'_x f'_y \cov(\overline X,\overline Y) 
		     + 2 f'_y f'_{z} \cov(\overline Y, \overline {Z}) + 2 f'_z f'_{x} \cov(\overline {Z}, \overline X)
\notag 		     \\
		     & \hspace{1em} 
		     +  \mathcal O(N^{-2}) \notag \\ 
		      &= \frac{1}{N} \Var_{q_{\theta}}[\blacktriangledown_\theta w] 
		      + \frac{\E_{q_{\theta}}[\blacktriangledown_\theta w]^2}{N } \Var_{q_{\theta}}[w] 
\notag \\
		      & \hspace{1em}
		      		      - 2 \frac {\E_{q_{\theta}}[\blacktriangledown_\theta w]} {N } \cov_{q_{\theta}}[\blacktriangledown_\theta w, w]
		      +  \mathcal O(N^{-2}).
		      \label{eq:VarianceZPathPQ}
\end{align}
We see that PathPQ and ZPathPQ have the same leading order variance (compare \eqref{eq:VariancePathPQ} and 
 \eqref{eq:VarianceZPathPQ}).
 This is because the difference  $f^{(\chi^2)}(\overline Y, \overline{ Z})$ between the two estimators is  $O_p(N^{-1})$ and thus its contribution to the variance is  $O(N^{-2})$ .

\subsection{Summary}
Our analysis revealed that
\begin{itemize}
\item
Both of the PathPQ and the ZPathPQ estimators have the biases,  \eqref{eq:biasPathPQ} and \eqref{eq:biasZPathPQ}, that are of $N^{-1}$ times smaller order than the true gradient.  
However, when the model distribution $q_{\theta}$ approaches to the target distribution $p$ in the converging phase of training, the leading order bias vanishes for PathPQ, while it stays for ZPathPQ.  Therefore, we can say that PathPQ has smaller bias than ZPathPQ in the converging phase.

\item
The PathPQ and ZPathPQ have the same leading order variances, \eqref{eq:VariancePathPQ} and \eqref{eq:VarianceZPathPQ}, which is $N^{-1}$ times smaller (hence the standard deviation is $N^{-1/2}$ times smaller) order than the sample gradients . 

\item For both estimators, the estimation error is dominated by the standard deviation, and therefore, we suppose that the advantage of PathPQ in terms of the bias in the converging phase does not have a large effect on the training performance.
\end{itemize}
We conclude that both estimators should perform similarly in the converging phase.

\section{
Initial Training Phase Behavior of Path Gradient Estimators of 
Forward KL Divergence
}
\label{app:initial-phase}
Weight degeneracy \cite{bugallo2017adaptive}  can become a serious issue for Importance Sampling.
Weight degeneracy is the phenomenon of only a few importance weights taking high values, while the other degenerate weights take negligible values.
At the start of training  in high-dimensional problem settings, it is often the case in practice that only a singular sample is non-degenerate. Here we will show this phenomenon is problematic for the ZPathPQ estimator. \\
In order to analyze this initial training regime theoretically, we assume without loss of generality that the $N$-th sample is singular, i.e., 
\begin{align}
    \frac{p(x_i)}{p(x_N)} = \mathcal{O}(\epsilon)\,, &&
        \frac{\|\nabla_{x_i} p(x_i)\|}{p(x_N)} = \mathcal{O}(\epsilon)
    \qquad \mbox{ for }i = 1, \ldots, N-1,
        \label{eq:SingularityAssumptionP}
\end{align}
and
\begin{align}
    q_\theta(x_i) = \mathcal{O}(1) \,, && \|\nabla_x q_\theta(x_i) \|= \mathcal{O}(1) \qquad \mbox{ for }i = 1, \ldots, N,
    \label{eq:SingularityAssumptionQ}
\end{align}
for small $\epsilon > 0$.
The former assumption \eqref{eq:SingularityAssumptionP} comes from the fact that most interesting densities in physics have an exponential fall-off around their modes,
while the latter assumption \eqref{eq:SingularityAssumptionQ} comes from the fact that all samples $\{x_{i}\}_{i=1}^N$ are generated from the sampler $q_{\theta}$.  
We also assume that $\epsilon \ll N^{-1}, d$, and ignore the scaling w.r.t. $N$ and $d$.

Let $\tau = p(x_N) $.%
\footnote{Note that it is assumed that $\tau = O(1)$ in Section~\ref{sec:theorytrainingphases} for simplicity.}
Then, the assumptions \eqref{eq:SingularityAssumptionP} and \eqref{eq:SingularityAssumptionQ} lead to
\begin{align*}
\nabla_{x_i} \frac{p({x_i})}{q({x_i})} &= \nabla_{x_i}( p({x_i}) )\frac 1 {q({x_i})} - \frac{p({x_i})}{q({x_i})^2}\nabla_{x_i} q({x_i}) \\
&= \mathcal O(\tau \epsilon) \frac 1 {q({x_i})} - \mathcal O(\tau  \epsilon)\nabla_{x_i} q({x_i}) \\
&= \mathcal O (\tau  \epsilon)
\end{align*}
for $i = 1, \ldots, N-1$.
Using the mild assumption that $\frac{\partial x}{\partial \theta} $ does not diverge,
the corresponding path-wise gradient is in the same order:
\begin{align}
    \blacktriangledown_\theta w(x_i) &= \frac{\partial w(x_i)}{\partial x_i}\frac{\partial x_i}{\partial \theta} = \mathcal O(\tau \epsilon).
\end{align}

\paragraph{PathPQ}
Under the singularity assumptions, \eqref{eq:SingularityAssumptionP} and \eqref{eq:SingularityAssumptionQ},
the PathPQ estimator \eqref{eq:PATHPQEstimatorApp} can be evaluated as
\begin{align*}
\text{PathPQ}_N &=
\sum_{i=1}^N \frac{1}{\sum_{j=1}^N {w}_j} \blacktriangledown_\theta {w}_i \\
&= \frac{1}{\sum_{j=1}^N {w}_j} \left(\sum_{i=1}^{N-1} \blacktriangledown_\theta {w}_i + \blacktriangledown_\theta {w}_N \right) \\
&= \frac{1}{w_N} (1 - \mathcal O(\epsilon)) (\blacktriangledown_\theta {w}_N + \mathcal O(\epsilon)) \\
&= \frac{\blacktriangledown_\theta w_N}{w_N} + \mathcal O(\epsilon).
\end{align*}
Therefore, if $\|\nabla_{x_N} p(x_N)\| = O(\tau) $ and hence $\frac{\blacktriangledown_\theta w_N}{w_N} = O(1)$,
the gradient direction is dominated by the singular ($N$-th) sample.
Interestingly, the dominating term for the PathPQ gradient is proportional to the dominating term for the PathQP gradient:
\begin{align*}
\text{PathQP}_N
&=
    \frac{1}{N} \sum_{i=1}^N \blacktriangledown_\theta w_i = \frac{1}{N }\blacktriangledown w_N + \mathcal{O}(\tau \epsilon).
\end{align*}
This implies that the PathPQ and the PathQP gradient behave similarly in the initial training phase, 
and could point to an explanation to the observation of \citet{geffner2021empirical} that their path-wise gradient estimators for alpha-divergences -- such as the forward KL -- seem to optimize the reverse KL in a high dimensional setting.

\paragraph{ZPathPQ}
Under the same singularity assumptions, \eqref{eq:SingularityAssumptionP} and \eqref{eq:SingularityAssumptionQ},
the ZPathPQ estimator \eqref{eq:ZPATHPQEstimatorApp} on the other hand does not have an order one contribution:
\begin{align*}
\text{ZPathPQ}_N
&= \sum_{i=1}^N \left(\frac{1}{\sum_{j=1}^N {w}_j} - \frac{w_i}{\left(\sum_{j=1}^N w_j \right)^2} \right)
\blacktriangledown_\theta {w}_i\\
&= \left(\frac{1}{\sum_{j=1}^N {w}_j} - \frac{w_N}{\left(\sum_{j=1}^N w_j \right)^2} \right)
\blacktriangledown_\theta {w}_N + \mathcal O(\epsilon) \\
&= \left(\frac{1}{w_N} - \frac{w_N}{w_N^2} \right)
\blacktriangledown_\theta {w}_N + \mathcal O(\epsilon) \\
&= \mathcal O(\epsilon).
\end{align*}
The ZPathPQ estimator is thus expected to struggle in the initial training phase with its weak gradient signal. Figure~\ref{fig:DReG-Initial-Phase} empirically validates this behavior.
This might also explain why the ZPathPQ estimator shows instable behavior in optimizing VAEs (see e.g. experiments on structured MNIST in \citet{tucker2018doubly}).\footnote{The reference refers to the PathPQ and ZPathPQ estimators for the loss of the variational encoder as IWAE-STL and RWS-DReG respectively.}

\section{Experimental details}
\label{app:exp-details}
\subsection{Double-Well}
For the quantum mechanical particle in the double-well potential, we trained a flow with RealNVP couplings which uses tanh activation, 8 coupling layers with 3 fully connected layers with a width of 200 neurons each.
The batch size was 4000. We used ADAM with an initial learning rate of 5$\times 10^{-5}$, $\beta=(0.9, 0.999)$.
The learning rate was decreased using a plateau learning rate schedule with a patience of 3000 to a minimum of $10^{-7}$. 
The base distribution for the normalizing flow was a univariate normal distribution with a standard deviation of 10.
The baselines estimators were run for 200k and 170k epochs, while the path gradient estimators were run for 100k epochs,
this ensured that the walltime duration for training the path-wise gradient estimators did not exceed the duration of training the baselines.
Training was done on NVidia P100 GPUs. 

\subsection{Estimating the forward ESS}
For the quantum mechanical particle in the symmetric double-well, we follow \cite{nicoli2021estimation}. Ten Hamiltonian Markov Chains (HMCs) were run with 100k steps, 50 sub-steps and an overrelax frequency of 10, totalling in 1 millions samples. We used 10k equilibrating steps.
An overrelax step mirrors the sample around zero. Due to the symmetry of the Double-Well, both the original and the mirrored sample have the same probability. Thus a MC step would always be accepted.

\subsection{Full results}
\begin{table}[ht]
    \centering
\begin{tabular}{l|r||lll|ll}
	{} & {$d$} & {ReinfPQ} & {ZPathPQ} & {PathPQ} & {RepQP} & {PathQP} \\
	\hline
    \multirow[c]{5}{*}{FW ESS} 
    & 8 & 0.99 $\pm$ .01 & \textbf{1.00} $\pm$ .00 & \textbf{1.00} $\pm$ .00 & 0.57 $\pm$ .20 & \textbf{1.00} $\pm$ .00 \\
    & 16 & 0.79 $\pm$ .18 & \textbf{0.99} $\pm$ .00 & \textbf{0.99} $\pm$ .00 & 0.79 $\pm$ .18 & 0.69 $\pm$ .18 \\
    & 32 & 0.45 $\pm$ .18 & \textbf{0.97} $\pm$ .01 & \textbf{0.90} $\pm$ .05 & 0.06 $\pm$ .02 & 0.20 $\pm$ .17 \\
    & 64 & 0.00 $\pm$ .00 &\textbf{0.17} $\pm$ .15 & \textbf{0.04} $\pm$ .03 & 0.00 $\pm$ .00 & 0.00 $\pm$ .00 \\
    \hline
    \hline
    \multirow[c]{5}{*}{Rev ESS} 
    & 8 & 1.00 $\pm$ .00 & \textbf{1.00} $\pm$ .00 & \textbf{1.00} $\pm$ .00 & \textbf{1.00} $\pm$ .00 & \textbf{1.00} $\pm$ .00 \\
    & 16 & 0.99 $\pm$ .00 & \textbf{0.99} $\pm$ .00 & \textbf{0.99} $\pm$ .00 & 0.99 $\pm$ .00 & \textbf{0.99} $\pm$ .00 \\
    & 32 & 0.95 $\pm$ .01 & \textbf{0.98} $\pm$ .00 & 0.98 $\pm$ .00 & 0.85 $\pm$ .05 & \textbf{0.98} $\pm$ .00 \\
    & 64 & 0.48 $\pm$ .13 & \textbf{0.90} $\pm$ .01 & \textbf{0.69} $\pm$ .14 & \textbf{0.60} $\pm$ .18 & 0.53 $\pm$ .13 \\
    \end{tabular}

    \caption{Results of training a RealNVP for $m_0$: 2.75}
    \label{tab:m02.75}
\end{table}

\begin{figure}[ht]
    \centering
    $m_0: 2.75$ \\
    \includegraphics[width=0.48 \textwidth]{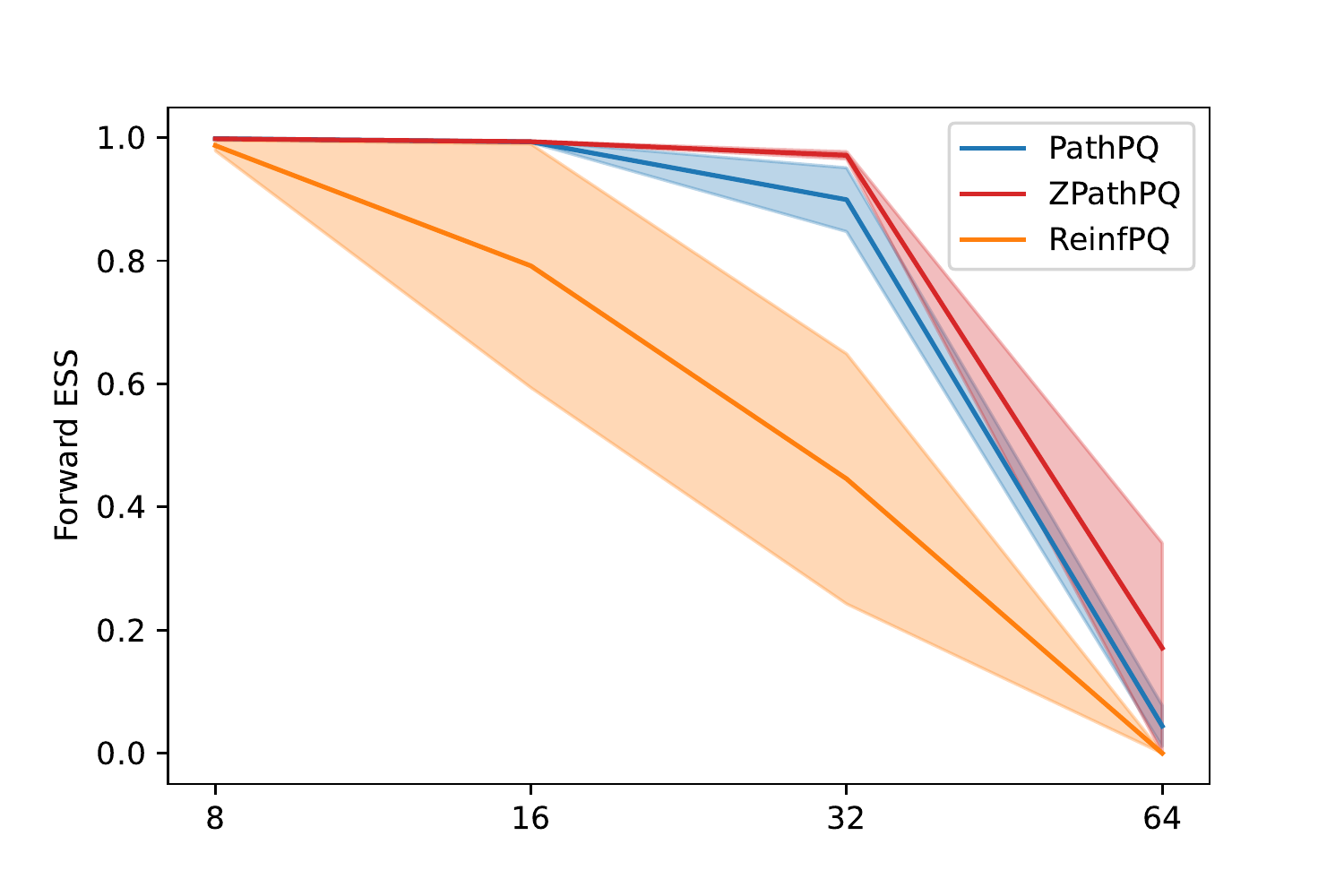}
    \includegraphics[width=0.48 \textwidth]{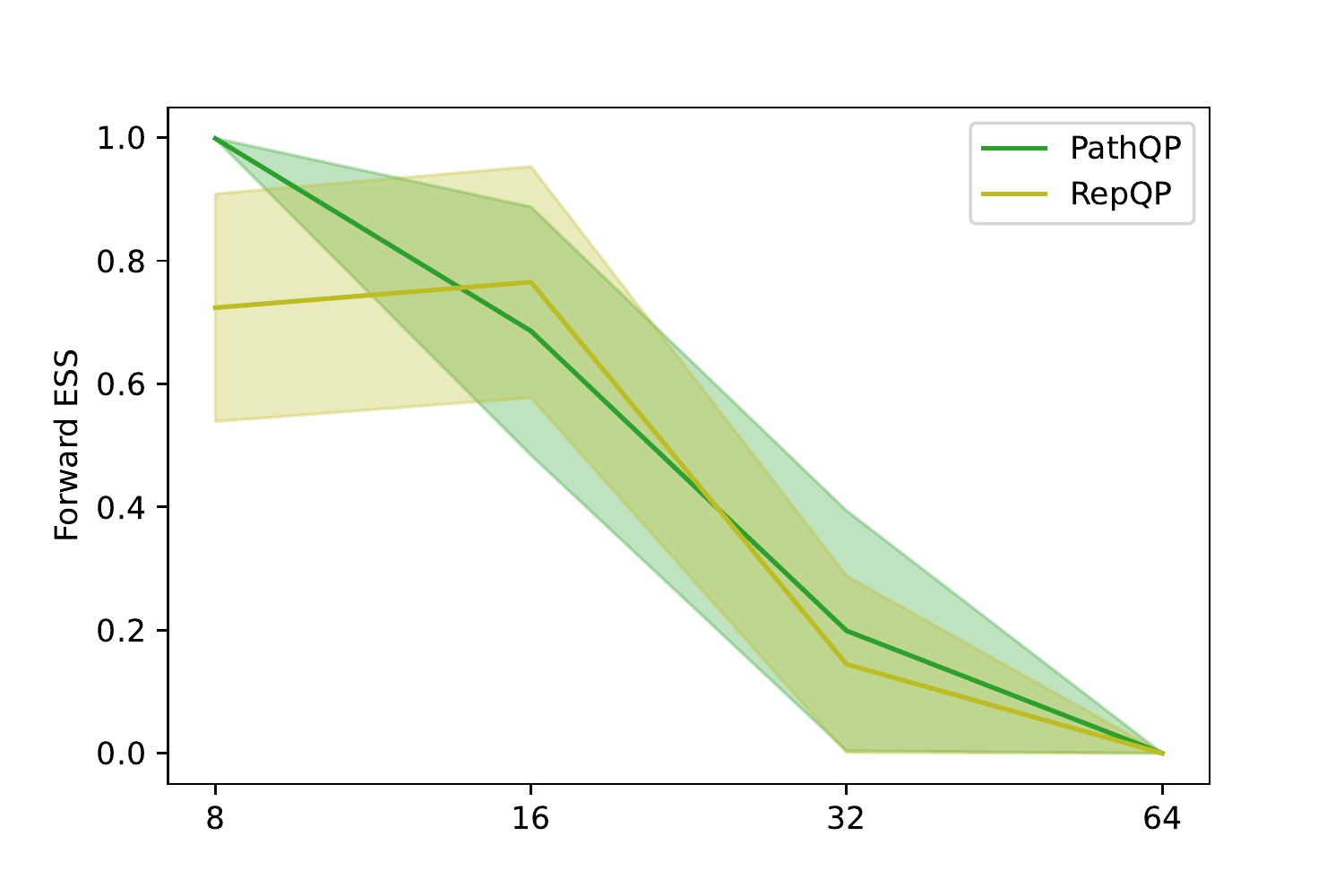} \\
       $m_0: 3.0$ \\
    \includegraphics[width=0.48 \textwidth]{figs/ForwardESS/figure8.pdf}
    \includegraphics[width=0.48 \textwidth]{figs/ForwardESS/figure9.pdf} \\
        $m_0: 3.25$ \\
    \includegraphics[width=0.48 \textwidth]{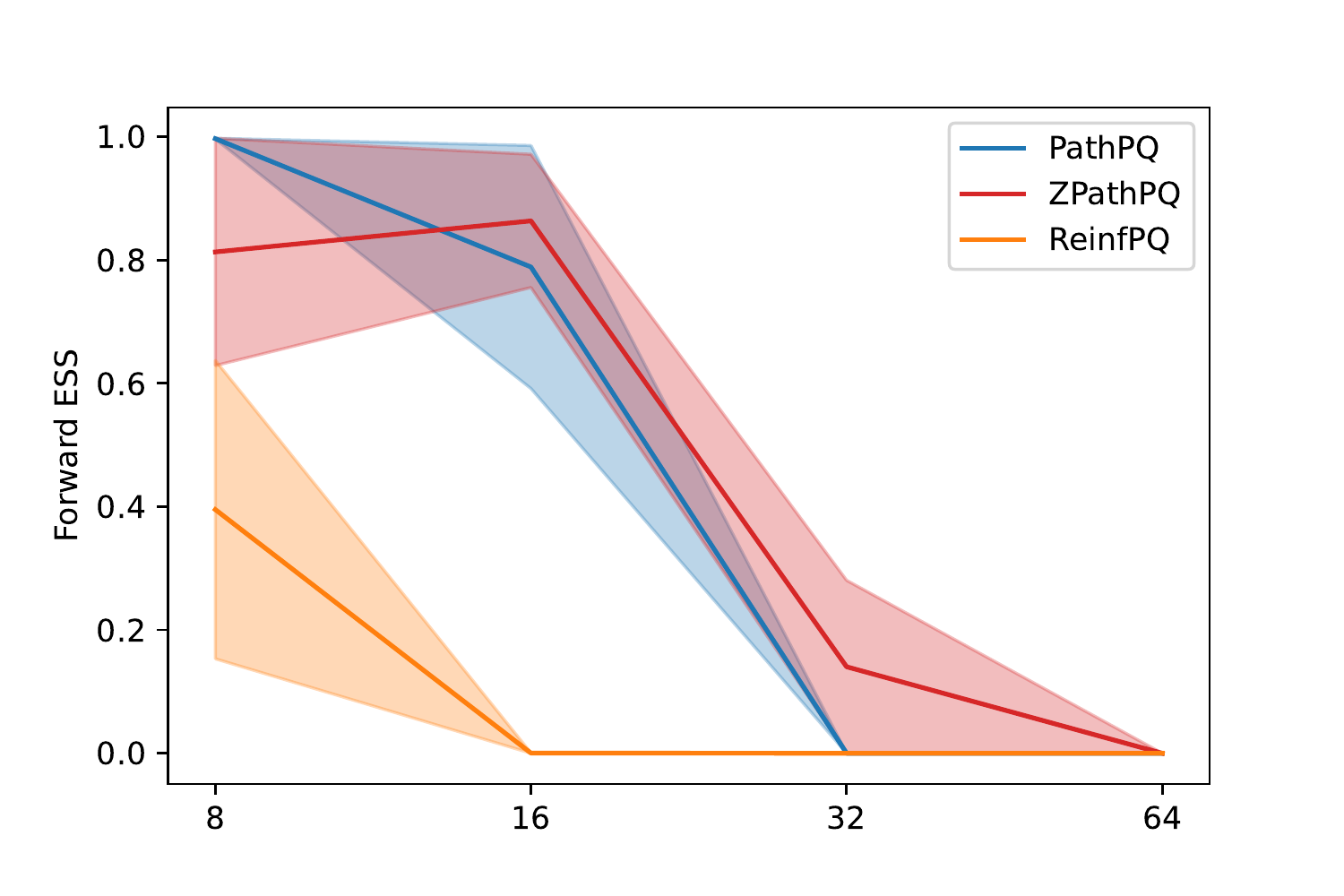}
    \includegraphics[width=0.48 \textwidth]{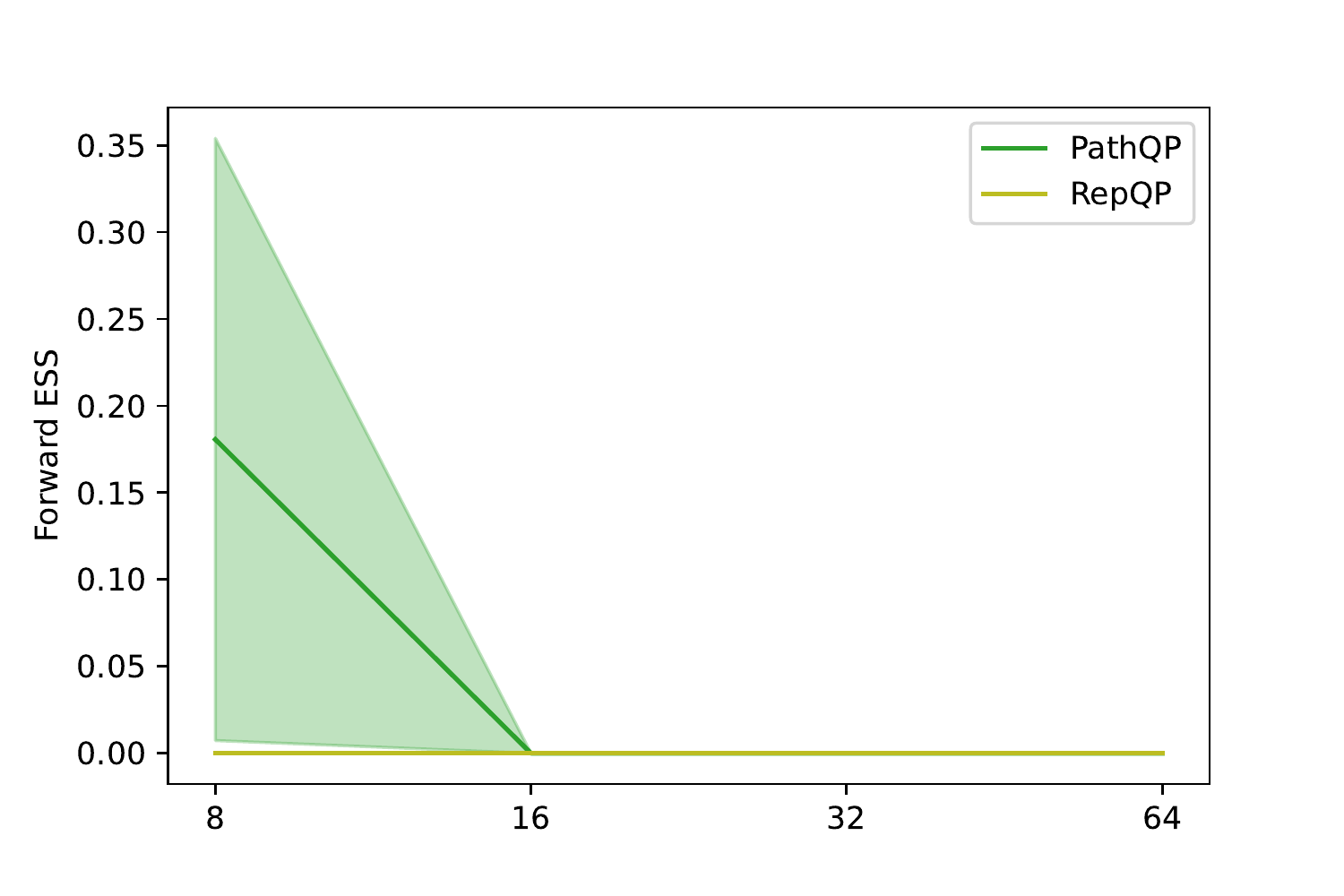} \\
    \caption{Forward ESS of trained RealNVP by varying over gradient estimators, dimensionality and separation parameter $m_0$}
\end{figure}

\end{document}

%% file: math_commands.tex
\usepackage{amsmath,amsfonts,bm}

\def\Secref#1{Section~\ref{#1}}

\def\eqref#1{(\ref{#1})}
\def\Eqref#1{Equation~\ref{#1}}

\def\1{\bm{1}}

\DeclareMathAlphabet{\mathsfit}{\encodingdefault}{\sfdefault}{m}{sl}
\SetMathAlphabet{\mathsfit}{bold}{\encodingdefault}{\sfdefault}{bx}{n}

\newcommand{\E}{\mathbb{E}}

\newcommand{\Var}{\mathrm{Var}}

\newcommand{\Cov}{\mathrm{Cov}}
